\pdfoutput=1

\documentclass[11pt]{article}

\usepackage[preprint]{acl}

\usepackage{times}
\usepackage{latexsym}

\usepackage[T1]{fontenc}

\usepackage[utf8]{inputenc}

\usepackage{microtype}

\usepackage{inconsolata}

\usepackage{graphicx}

\usepackage[all]{tcolorbox}
\usepackage{listings}
\usepackage{caption}

\usepackage{booktabs}
\usepackage{multirow}
\usepackage{array}
\usepackage{wrapfig}

\makeatletter
\renewcommand{\@makefnmark}{\textcolor{black}{*}}
\makeatother

\lstdefinelanguage{none}{}

\newtcblisting{tcbverbatim}[2][]{ %
  blank,
  borderline={1pt}{-2pt},
  listing only,
  breakable,
  enhanced jigsaw,
  colback=white,
  left=10pt,
  right=10pt,
  toptitle=5pt,
  title={#1},
  fonttitle=\small\bfseries, %
  coltitle=black,
  colbacktitle=gray!10,
  listing options={
    language=none,
    escapeinside={(*@}{@*)}, %
    basicstyle=\ttfamily\small\parindent=0pt,
    columns=flexible,
    breaklines=true,
    breakatwhitespace=true,
    breakautoindent=false,
    breakindent=0pt,
    #2 %
  }
}

\newcommand{\blob}[2]{
    \vcenter{\hbox{\begin{tikzpicture}
    \node[rounded corners, fill=#1] {\sffamily\scriptsize{\color{white}{#2}}};
    \end{tikzpicture}}}
}
\definecolor{left}{rgb}{0.94, 0.5, 0.5}
\definecolor{right}{rgb}{0.53, 0.81, 0.98}
\definecolor{target}{rgb}{0.87, 0.46, 0.0}
\definecolor{chosen}{rgb}{0.29, 0.59, 0.82}
\definecolor{10}{rgb}{0.899, 0.531, 0.548}
\definecolor{115}{rgb}{0.893, 0.536, 0.555}
\definecolor{20}{rgb}{0.858, 0.562, 0.596}
\definecolor{252}{rgb}{0.837, 0.578, 0.621}
\definecolor{454}{rgb}{0.754, 0.641, 0.718}
\definecolor{50}{rgb}{0.735, 0.655, 0.74}
\definecolor{70}{rgb}{0.653, 0.717, 0.836}
\definecolor{763}{rgb}{0.627, 0.737, 0.867}
\definecolor{935}{rgb}{0.557, 0.78, 0.949}

\title{On the Same Wavelength? Evaluating Pragmatic Reasoning \\ in Language Models across Broad Concepts}

\author{Linlu Qiu\thanks{Equal contributions.}$^{1}$\quad Cedegao E. Zhang*$^{1,2}$ \\
\bf Joshua B. Tenenbaum$^{1,2}$ \quad Yoon Kim$^1$ \quad Roger P.\ Levy$^2$ \\
\texttt{\{linluqiu, cedzhang\}@mit.edu} \\
$^1$MIT CSAIL \quad $^2$MIT BCS 
}

\begin{document}
\maketitle
\makeatletter
\renewcommand{\@makefnmark}{\hbox{\@textsuperscript{\normalfont\@thefnmark}}}
\makeatother
\begin{abstract}
Language use is shaped by pragmatics---i.e., reasoning about communicative goals and norms in context. As language models (LMs) are increasingly used as conversational agents, it becomes ever more important to understand their pragmatic reasoning abilities. We propose an evaluation framework derived from \textit{Wavelength}, a popular communication game where a speaker and a listener communicate about a broad range of concepts in a granular manner. We study a range of LMs on both language comprehension and language production using direct and Chain-of-Thought (CoT) prompting, and further explore a Rational Speech Act (RSA) approach to incorporating Bayesian pragmatic reasoning into LM inference. We find that state-of-the-art LMs, but not smaller ones, achieve strong performance on language comprehension, obtaining similar-to-human accuracy and exhibiting high correlations with human judgments even without CoT prompting or RSA. On language production, CoT can outperform direct prompting, and using RSA provides significant improvements over both approaches. Our study helps identify the strengths and limitations in LMs' pragmatic reasoning abilities and demonstrates the potential for improving them with RSA, opening up future avenues for understanding conceptual representation, language understanding, and social reasoning in LMs and humans.\footnote{Code and data are available at \url{https://github.com/linlu-qiu/wavelength-eval}.}
\end{abstract}

\section{Introduction}

Human communication occurs \emph{in context}, undergirded by shared goals, norms, and situational cues that shape communication beyond literal meanings of utterances. 
Within linguistics and cognitive science, \emph{pragmatics}  provides a  broad framework for studying how speakers and listeners use and interpret language in context \citep{grice1975logic,levinson1983pragmatics}. Pragmatic communication broadly supposes that given a shared context and communicative goals between speakers and listeners, the speaker chooses utterances to convey specific meanings, while the listener interprets the speaker's intended meanings by assuming that the speaker is being cooperative and informative. This interplay can lead to rich and nuanced phenomena observed in human communication, such as implicature, ambiguity, vagueness, hyperbole, and more \citep{wittgenstein1953philosophical,austin1962things, sperber1986relevance, clark1996using}.

\begin{figure}
    \centering
    \includegraphics[width=0.95\linewidth]{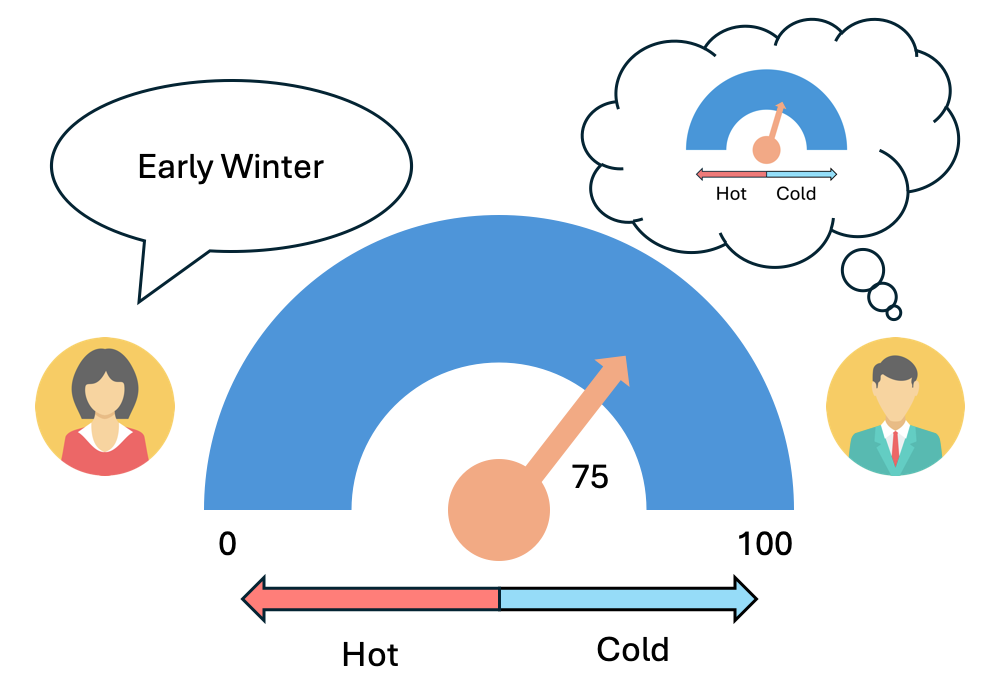}
    \caption{A visual illustration of our experimental setup based on the Wavelength game. In the production task, the speaker (left) is asked to generate a clue (``Early Winter'') given the pair of opposite concepts (``Hot'' and ``Cold'') and target value (75). In the comprehension task, the listener (right) is asked to make a guess about the target value given the concepts and clue.}
    \label{fig:teaser}
\end{figure}

State-of-the-art language models (LMs) have made significant progress on language understanding and generation, and  have now advanced to the point where they are being used actively as conversational agents by millions of people \citep{wu2023brief, xi2025rise}. Assessing the pragmatic reasoning abilities of LMs is thus of both theoretical and practical interest. 
Existing studies of LMs' pragmatic reasoning  generally fall into two settings. The first setting adopts a benchmark approach and evaluates LMs on test materials that target various pragmatic phenomena such as presupposition and implicature \citep{hu2022fine, ruis2023goldilocks, sravanthi2024pub}. These studies generally find that large LMs can achieve high accuracy and match human error patterns to an extent. The second setting instead asks LMs to act as listeners and/or speakers in communication settings or games and compares the behaviors of LMs against those of human subjects \citep{jian2024llms,tsvilodub2025non, ma2025vision}. In this setting, even frontier LMs may not match human performance and demonstrate gaps in their pragmatic reasoning.

We study the  pragmatic reasoning capabilities of LMs in the second setting through \emph{Wavelength},\footnote{\url{https://www.wavelength.zone}} a popular verbal communication game in which one player (listener) tries to guess a hidden number (between 0 and 100) on a scale between two opposite concepts (e.g., ``Hot'' and ``Cold'') based on a clue given by another player (speaker), who is trying to communicate the hidden number (see Fig.~\ref{fig:teaser} for an example). While conceptually simple, Wavelength captures several core phenomena crucial for effective pragmatic communication. These include the pursuit of shared goals (measured by the guesser's success), representation of graded meanings (measured on a continuous scale), and application of world knowledge across a diverse set of concepts.

Importantly, Wavelength supports open-ended reasoning about a broad range of words and phrases, but does so  in a controlled setting---a property shared by many good games that are useful for studying the mind \citep{allen2024using}. It allows us to collect granular, graded judgments (i.e., numeric values on a continuous scale) from both LMs and humans and ground our quantitative analysis in human distributional data \citep[cf.][]{ying2025benchmarking}. In this setting, we study both the comprehension and production sides of pragmatic reasoning, as both are fundamental to conversational agents. We further experiment with incorporating Rational Speech Act \citep[RSA,][]{goodman2016pragmatic}, a leading computational pragmatics framework, into LMs and assess whether RSA improves model performance or fit to human data. 
 
We test different families of LMs on this dataset and find that model performance systematically increases with model size. On the comprehension task, state-of-the-art models perform at near-human accuracy and show strong correlations with human judgments even with just direct prompting. However, all LMs show significant divergence from humans in terms of the \textit{distributions} of judgments, where we find that human samples are more varied. On the production task, larger models also perform strongly, but more interestingly, we find that  RSA-augmented LMs significantly improve upon  both direct and Chain-of-Thought (CoT) prompting. 

In sum, our main contributions in this paper are:

\begin{itemize}
    \item A novel dataset collected from human experiments for benchmarking LMs' pragmatic language comprehension and production abilities.
    \item An LM-based probabilistic inference method that leverages RSA for pragmatic reasoning.
    \item Evaluations of a range of LMs, finding that larger models perform well in terms of accuracy and human correlations on the comprehension task and LM-RSA reliably helps the production task.
\end{itemize}
Our study also sheds light on the nature of conceptual representation and language processing in humans and LMs, as we discuss at the end.

\begin{table*}[ht!]
\footnotesize
\centering
\scalebox{0.95}{
\begin{tabular}{@{}cccp{6cm}c>{\centering\arraybackslash}p{1.2cm}@{}}
\toprule
\multirow{2}{2cm}{\centering\textbf{Left Concept (0)}} & \multirow{2}{1cm}{\centering\textbf{Target Value}} & \multirow{2}{2cm}{\centering\textbf{Right Concept (100)}} & \multirow{2}{6cm}{\centering\textbf{Human-written Clues}} & \multirow{2}{1.4cm}{\centering\textbf{Chosen Clue}} & \multirow{2}{1cm}{\centering\textbf{Human Mean}} \\
& \\
\toprule
${\blob{left}{Deep thought}}$ & ${\blob{10}{10}}$ & ${\blob{right}{Shallow thought}}$ & Evolution, Solving complex problems, Chess, Einstein, Meditation, Quantum mechanics & ${\blob{115}{Solv. complex prob.}}$ & 11.5 \\
${\blob{left}{Hot}}$ & ${\blob{70}{70}}$ & ${\blob{right}{Cold}}$ & Coatless, Iced coffee, Refrigerator, Temperature, Colorado mountains, Early winter & ${\blob{763}{Early winter}}$ & 76.3 \\ 
${\blob{left}{Mental activity}}$ & ${\blob{right}{100}}$ & ${\blob{right}{Physical activity}}$ & Running, Sprint, Work out, Race, Gym workout, Marathon & ${\blob{935}{Sprint}}$ & 93.5 \\
${\blob{left}{Messy food}}$ & ${\blob{20}{20}}$ & ${\blob{right}{Clean food}}$ & Crawfish, Tacos, Spaghetti, Ribs, Stuffed burrito, Big Mac & ${\blob{252}{Big Mac}}$ & 25.2 \\
${\blob{left}{Sport}}$ & ${\blob{50}{50}}$ & ${\blob{right}{Game}}$ & Betting, Darts, Table tennis, FIFA, Mini golf & ${\blob{454}{Table tennis}}$ & 45.4 \\
\bottomrule
\end{tabular}%
}
\caption{Example human data. Colored chips represent building blocks of experimental stimuli, and the colors represent the spectrum between 0 and 100. For each production problem we collect human-written clues given the left concept, target value, and right concept. We use human judgments to choose the best clue for the comprehension experiment (chosen clue), and obtain final comprehension judgments (human mean) given the left concept, chosen clue, and right concept. More examples also appear in Table~\ref{tab:qualitative_clue}; all the concept pairs are presented in Appendix \ref{sec:concept_pairs}.}
\label{tab:examples}
\end{table*}

\section{Background and Related Work}
\label{sec:related_work}

\paragraph{Models of pragmatics.}

Pragmatics has long been a central subject in the philosophy and science of language \citep{sep-pragmatics}. Early theorists have emphasized the social nature of language and analyzed a wide range of interesting or puzzling phenomena in verbal communication that arise because of pragmatics \citep{searle1969speech, grice1975logic, clark1996using}. Formal theories have been developed to target some of them, yet those are not grounded in large-scale empirical studies \citep{Stalnaker1978-STAA-2, heim1982semantics, chierchia2012scalar, kamp2013discourse}. More recently, the Rational Speech Act (RSA) represents a promising framework in computational pragmatics that can make quantitative predictions about human pragmatic understanding and reasoning \citep{frank2012predicting, goodman2016pragmatic, degen2023rational}. At its core, RSA posits a cooperative speaker who selects utterances to maximize conversational utility, and a listener who interprets these utterances through Bayesian inference. It has been used to model a wide range of pragmatic phenomena, including hyperbole, vagueness, generics, and politeness \citep{kao2014nonliteral, lassiter2017adjectivalVI, tessler2019language, yoon2020polite}.

\paragraph{Pragmatics in language models.}

There is a growing body of work evaluating the pragmatics in LMs~\citep[see \citet{fried2023pragmatics} and \citet{ma2025pragmatics} for surveys]{zheng2021grice, li2023diplomat, ruis2023goldilocks, liu2024large, sravanthi2024pub, zhao2025comparing}. Among them, \citet{hu2022fine} is related to our work in that it systematically compares LM and human judgments yet differs in that it utilizes multiple-choice questions rather than graded judgments. Our work shares similarity with \citet{lipkin2023evaluating}, which collects graded judgments (on a likert scale) and compares LM and human distributions. 
But that work only focuses on one domain and studies one LM \cite[OpenAI Codex,][]{chen2021evaluating}. 
Notably, concurrent works \citet{tsvilodub2025non} and \citet{spinoso-dipiano2025rsa} also pursue the direction of incorporating LMs and RSA and ground it in rigorous evaluations, but they focus only on the comprehension side and on a few specific pragmatic phenomena (both on hyperbole and pragmatic halo effects; the latter additionally on irony). On the production side, \citet{jian2024llms} evaluates the LM using a reference game and find they are not good pragmatic speakers, consistent with our results. \citet{junker2025multimodal} and \citet{ma2025vision} present benchmarks on multimodal pragmatic expression generation, but they mostly focus on evaluating the vision-language model performance. 
\citet{murthy2025inside} also uses RSA model to study LMs, but focuses on interpret value trade-offs in LMs.

\section{The WavelengthEval Dataset}
\label{sec:data}

As previously introduced, the Wavelength game captures core features of communication that require pragmatic reasoning. We use it to design a human data collection procedure and create a resulting benchmark for LM pragmatic reasoning evaluation. 
The core concepts are a pair of opposite or contrasting concepts (spatially \textit{left} and \textit{right}, representing the two extremes and a spectrum in between), a \textit{target value} (between 0 and 100, inclusive), and a \textit{clue} (a single word or a short phrase). We conduct experiments on both the \textit{production} and \textit{comprehension} tasks: for production, the left and right concepts and a target value are given, and the goal is to generate a clue that best communicates the target to another competent language user (maximizing their chance of guessing the value correctly); for comprehension, analogously, the concepts and a clue are given, and the goal is to make the best guess at the intended target value. 

These tasks have several properties that make them ideal for human and LM evaluations: (1) the left-right concept pairs are broad and diverse (see examples in Table \ref{tab:examples}), including both concrete and abstract ones. (2) The choice for clues is open-ended (requiring agents to think over a large set of possible options) but still controlled (only at most a few words are allowed). (3) The target values are precise and graded, making it much harder to blindly guess. (4) The distribution of guesses reflects agents' uncertainty about the problems. These properties match or go beyond well-established paradigms in cognitive science \citep{griffiths2024bayesian}. From a linguistic theory perspective, our test items manifest a range of pragmatic phenomena; examples include scalar adjectives and vagueness (``How cold is cold?''), ad-hoc concepts (using ``Einstein'' to refer to ``deep thought''), and comparison class inference (``Are Jeans considered long---among pants versus all clothing items?'') \cite{kennedy2007vagueness, barsalou1983ad, tessler2022warm}.

We note that for evaluation purposes, people's comprehension task performance can be measured against the true underlying target value. However, there is no ground truth for judging people's clue generation---the gold standard is to collect human comprehension judgments. In other words, the ground truth evaluation of comprehension is much less costly than that of production. This asymmetry motivates a two-phase experimental design. In Phase 1, we collect a set of human-generated clues for each production problem. Then for each clue we collect a set of human comprehension judgments to filter out the best clue for each problem. The resulting clues form the stimuli for Phase 2, where we expand the set of human judgments for each comprehension problem. The data collection process is as follows, where more details about the human experiments can be found in Appendix~\ref{app:human_exp}.

\paragraph{Stimuli.} We first obtain the official set of pairs of concepts in Wavelength.\footnote{\url{https://boardgamegeek.com/thread/2387770/card-list}} There are over 200 pairs, and among them we manually choose 50 that are comparatively more commonsensical and less subjective as our stimuli. For each pair, we manually assign two target values that are multiples of 5 between 0 and 100, obtaining 100 problem instances in total. The overall distribution of target values across problems is roughly uniform.  

\paragraph{Participants.} We recruit 708 human participants in total from the Prolific platform \citep{palan2018prolific}. The human experiments are approved by the local university IRB. For all experiments we apply the following criteria: adult (default), living in USA, fluent in English, and approval rate 99\%+.

\paragraph{Outcome.} Our experimental process results in a dataset, WavelengthEval, that contains 50 unique pairs of concepts with 2 target values. For each left-right-value triple, we have an empirically determined high-quality human generated clue and correspondingly 40 human guesses.
We then use this dataset to evaluate pragmatic reasoning in LMs through both the comprehension and production tasks, beginning with the former.

\section{LM Comprehension}
\label{sec:comprehension}
Given two opposite concepts and a clue, the language comprehension task requires the listener to guess a value that best represents the clue. An ideal listener should predict a value that is close to the speaker's target. For both humans and LMs, we first estimate their distributions over the scale for each problem and then use the mean 
as their prediction.

\begin{figure*}[t]
    \centering
    \includegraphics[width=\linewidth]{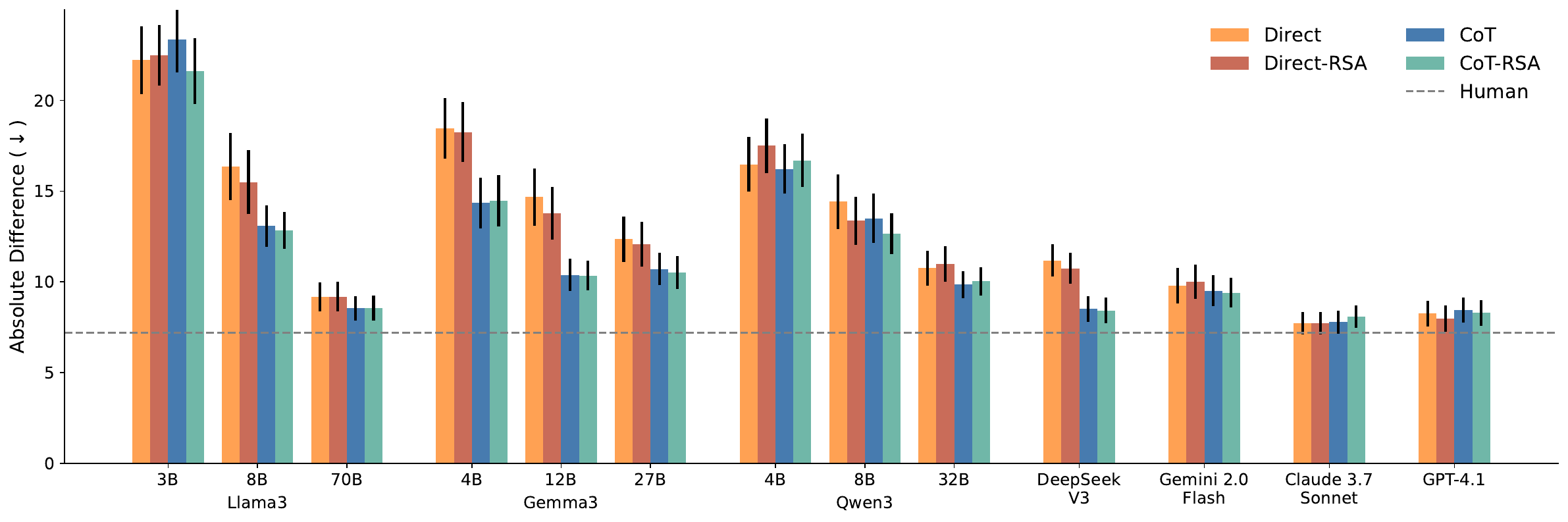}
    \vspace{-0.5cm}
    \caption{Absolute difference between the model's prediction and the target value. Error bars show standard error over each problem. The dashed line indicates human performance.}
    \label{fig:listener_diff}
    \vspace{-0.3cm}
\end{figure*}

\begin{figure*}[t]
    \centering
    \includegraphics[width=\linewidth]{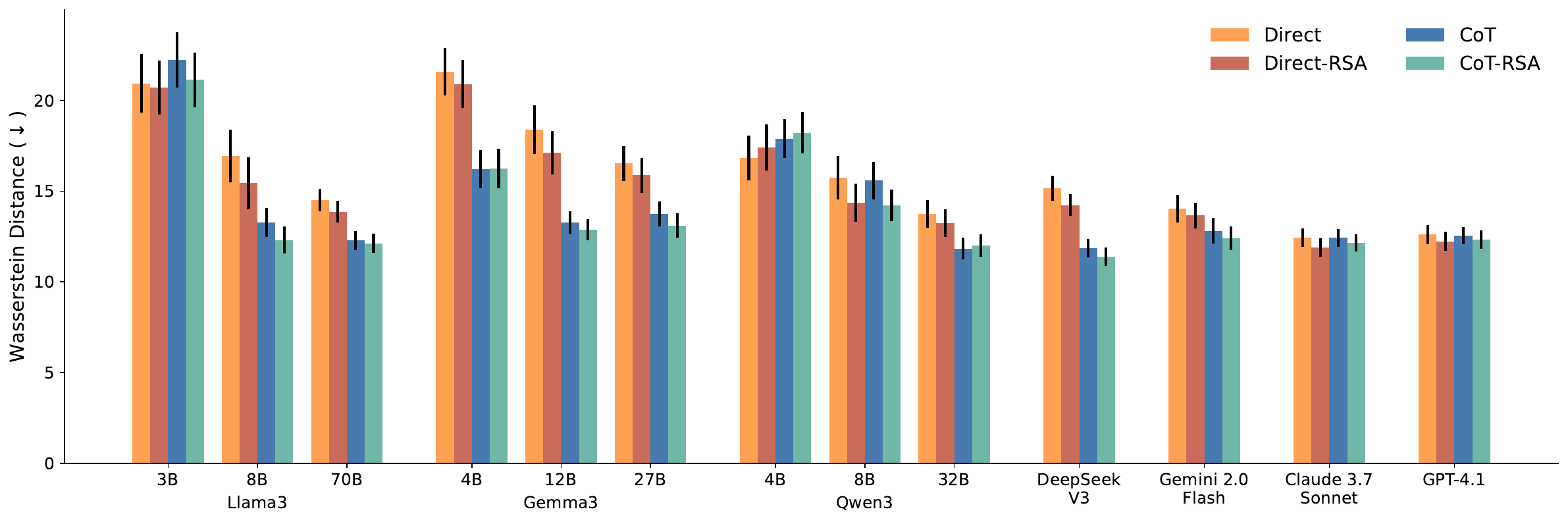}
    \vspace{-0.5cm}
    \caption{The Wasserstein distance between the model distribution and human distribution. Error bars show standard error over each problem.}
    \label{fig:listener_human_dist}
    \vspace{-0.4cm}
\end{figure*}

\subsection{Methods}

We use open-weights, ``instruct'' versions of LMs for most of our experiments as they provide models of various sizes and access to logits. We consider Llama3 (3.2 3B, 3.1 8B, 3.3 70B) \citep{llama3}, Gemma3  (4B, 12B, 27B)~\citep{gemma3}, Qwen3  (4B, 8B, 32B) \citep{yang2025qwen3}, and DeepSeek-V3~\citep{deepseek-v3} families of models. We also evaluate three representative API-only models: Gemini 2.0 Flash~\citep{gemini}, Claude 3.7 Sonnet~\citep{claude37sonnet}, and GPT-4.1~\citep{openai2025gpt41}.

\begin{figure*}[t]
    \centering
    \includegraphics[width=\linewidth]{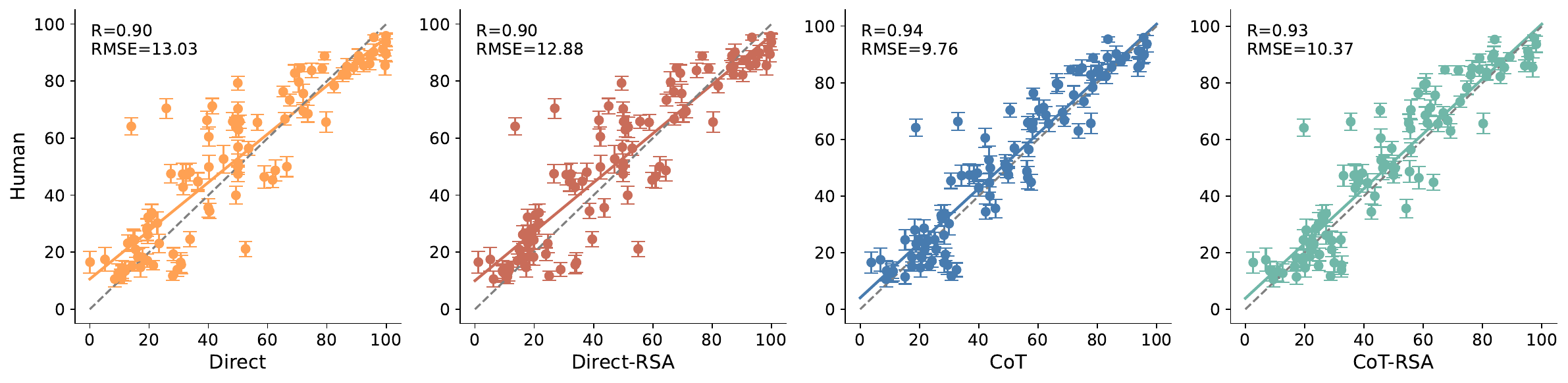}
    \caption{Correlations between model predictions and mean human judgments using Qwen3 32B (see Appendix~\ref{sec:app_full_results} for more results). We show Pearson correlations and root mean square standard error (RMSE). Error bars show standard error over 40 human participants.}
    \label{fig:corr}
\vspace{-3mm}
\end{figure*}

We evaluate the LM using three different methods. The first one is \textit{direct} prompting, where we obtain the LM distribution using a prompt that asks it to output responses without intermediate tokens (see Appendix~\ref{sec:app_prompts} for prompts). For open-weights models, we approximate this distribution by scoring the next-token probability of all possible values along the scale in increments of 5. For API-only models, we use 32  samples from the LM to estimate this distribution. The second method is \textit{zero-shot CoT} prompting \cite{kojima2022large}, where we instruct the model to think step-by-step before generating the final answer. We similarly use 32 samples to approximate this distribution.

The third method, \textit{LM-RSA}, incorporates probabilistic inference into the LM based on the RSA framework, which views communication as recursive reasoning between a speaker and a listener. We start from the literal listener $L_0$ (originally representing semantic understanding without pragmatic strengthening), which here is implemented by an LM using either direct or CoT prompting (\textit{Direct-RSA} or \textit{CoT-RSA}). This is based on the assumption that the LM has already acquired the basic meaning of an utterance.

In other words, we define
\[
P_{L_0}(s \mid u) \propto [\![ u ]\!](s) \cdot P(s) \approx P_{LM}(s \mid u) \cdot P(s),
\] 
where $s$ (state) refers to possible target values, $u$ (utterance) refers to possible clues, $[\![\cdot]\!]$ refers to a semantic interpretation function \citep{Heim1998-HEISIG}, and $P(\cdot)$ refers to the prior over states. Assuming a uniform distribution over all states, equal costs for all utterances, and the softmax rationality parameter $\alpha=1$, the pragmatic speaker \( S_1 \) selects an utterance \( u \) given a state \( s \) based on:
\begin{align*}
P_{S_1}(u \mid s) &\propto \exp(\alpha \cdot U_{S_1}(u; s)) \\ &= \frac{P_{L_0}(s \mid u)}{\sum_{u' \in \mathcal{U}} P_{L_0}(s \mid u')},
\end{align*}

where \( \mathcal{U} \) denotes a set of possible utterances, $U_{S_1}$ denotes a utility function, which we define as $\log P_{L_0}(s \mid u)$. Finally, the pragmatic listener $L_1$ computes the state probabilities given the utterance by performing Bayesian inference over  $S_1$, 
\[
P_{L_1}(s \mid u) = \frac{P_{S_1}(u \mid s) \cdot P(s)}{\sum_{s' \in \mathcal{S}} P_{S_1}(u \mid s') \cdot P(s')},
\]
where \( \mathcal{S} \) denotes the set of possible states. This process requires defining the set of alternative utterances, which we use the LM to generate. 
Specifically, for each state \( s \), we directly sample 1 alternative utterance from the LM without CoT (see prompts in Appendix~\ref{sec:app_prompts}).\footnote{We explored sampling more alternatives but did not observe significant improvements.} Using the LM as an alternative generator allows us to collect a flexible set of utterances, a task that was challenging in the traditional RSA framework, particularly in our open-ended setting. In Appendix~\ref{sec:app_listener_v2}, we explore a variant of the RSA listener model, which performs similarly to this version.

\subsection{Evaluation Metrics}

We evaluate the LM from two perspectives. The first is task performance, which evaluates how the LM performs on the comprehension task that requires pragmatic reasoning. We measure this using the absolute difference between the model's prediction and the speaker's target value. Secondly, the comprehension task inherently involves uncertainty. Although individuals may differ on exact values, there is generally consensus on an approximate range. For example, values in the 60-80 range might all be considered acceptable for ``early winter'' on the ``Hot-Cold'' scale. To further evaluate whether the LM captures this uncertainty and maintains a distribution similar to human judgments, we compare the model's distribution with the human distribution using the Wasserstein distance. We also use the Pearson correlations between human and model mean judgments to measure their similarities.

\subsection{Results}

\begin{figure*}[t]
    \centering
    \includegraphics[width=\linewidth]{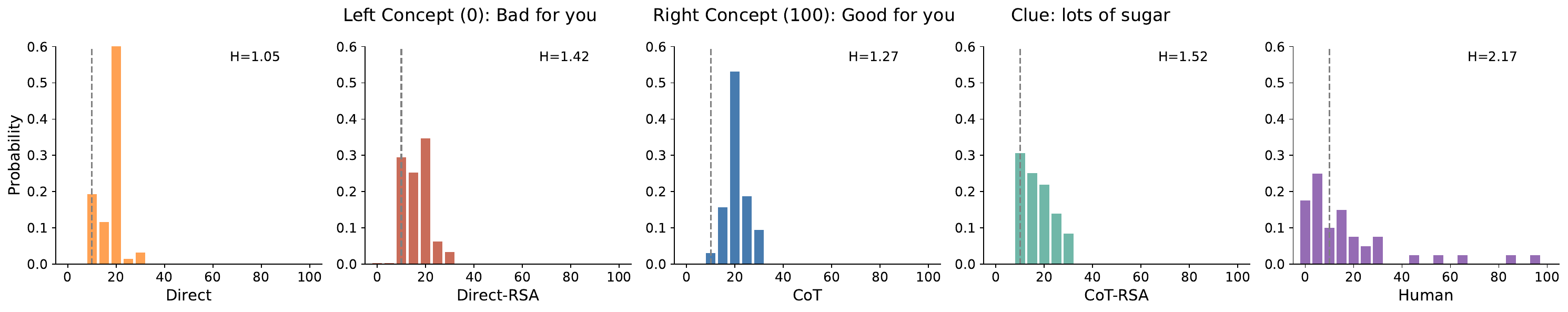}
    \includegraphics[width=\linewidth]{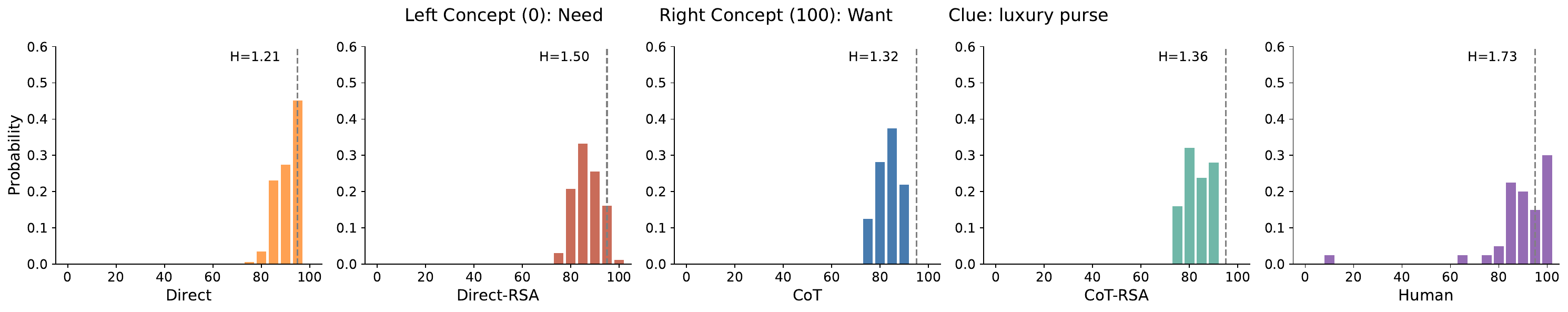}
\caption{Examples of the LM distribution and human distribution using Qwen3 32B. We show an example where RSA improves the base prompting methods on both task performance and humanlike-ness (top) and an example where RSA hurts (bottom). Dashed lines indicate target values. $H$ refers to the entropy of the distribution.}
\label{fig:listener_qualitative}
\vspace{-0.2cm}
\end{figure*}

\paragraph{Task performance.}
We show the absolute difference between the LM's prediction and target value in Fig.~\ref{fig:listener_diff}. Within each model family, the task performance improves as the model size increases, with the largest model achieving close to human performance. Across model families, the stronger API-only models generally achieve better performance than weaker models. Additionally, CoT improves performance over the direct approach, but RSA does not bring significant improvement.

\paragraph{Humanlike-ness.} 
We show  the Wasserstein distance between the model distribution and human distribution in Fig.~\ref{fig:listener_human_dist}.
Similar to the task performance, we find that CoT generally reduces the Wasserstein distance compared to the direct approach. Incorporating RSA helps when using the direct approach, but does not add further value to the CoT prompting. We calculate Pearson correlations between human judgments and the model predictions, as well as the root mean square standard error (RMSE). While smaller models generally show weaker correlation with human judgments, the best model from each model family is typically highly correlated with human judgments, achieving a correlation higher than 0.9 (see Appendix~\ref{sec:app_full_results} for details). Fig.~\ref{fig:corr} shows a breakdown of correlations between the model predictions and human judgments using Qwen3 32B, a representative open-weights model that achieves strong performance. We observe high correlations across different methods, demonstrating that the LM is able to capture mean human judgments to a large extent in this task.

\subsection{Analysis}

\paragraph{Do LMs and humans perform consistently across different target values?}
Our target values are roughly uniform from 0 to 100. However, performance on different values may be different for LMs or humans, as certain values might be easier to guess. We show the task performance breakdown by the target value in Fig.~\ref{fig:target_vs_diffs} (Appendix \ref{sec:app_results}). We find that humans perform worse towards the left extreme scale (0), but are generally consistent across different target values. The performance of LMs, however, shows larger variance across target values, especially for small models (e.g., Qwen3 8B, which performs significantly poorly near 0).

\paragraph{LM vs. human distribution.} We show examples of the LM and the human distribution in Fig.~\ref{fig:listener_qualitative}. The top one illustrates the case where using RSA reduces the absolute difference between the model's mean and the target, while also capturing the underlying uncertainty of the human distribution. Intuitively the model becomes less confident on an incorrect value. The bottom one demonstrates a failure case where using RSA hurts performance---the model becomes less confident at extreme values, plausibly because of the other alternatives it generates also assign extreme values.

\paragraph{Choices of alternatives.} One hypothesis for why RSA does not improve performance is that, as a listener, the LM tends to be over-confident, producing distributions that are generally concentrated and spiky. In such cases, RSA has limited effectiveness in reshaping the distribution.
An alternative hypothesis is that, as a simulated speaker, the LM does not generate good alternatives. We investigate which factor contributes more to the performance by varying the choices of alternatives. We observe that the LM performance is insensitive to the choices of alternatives—providing better or worse alternatives leads to similar performance (Appendix~\ref{sec:app_listener_alt}). Therefore, we hypothesize that RSA fails to further improve LM performance on the comprehension task because their concentrated and spiky listener distributions (see Fig.~\ref{fig:listener_qualitative} for examples and Appendix Fig.~\ref{fig:listener_entropy} for comparisons of entropy between the LM and human distributions), which limit the benefits of Bayesian re-normalization, the core mechanism of the RSA framework.

\paragraph{Qualitative example.} The other potential explanation for why we do not observe significant improvement when using RSA is that the LMs already perform pragmatic inference, either implicitly or explicitly with CoT. We qualitatively examine the reasoning chains of the models and find that in many cases, they already perform RSA-style inference by reasoning about the alternatives, which partially explains the improved performance of CoT compared to the direct approach. We show an example output of Qwen3 32B model in Appendix~\ref{sec:app_qualitative}.\footnote{This only applies to CoT prompting, where we have access to the model's reasoning chains. It is still possible that the model performs similar inference implicitly in the direct prompting case. We leave probing the model's internal mechanisms for future work.}

\begin{figure*}[t]
    \centering
    \includegraphics[width=\linewidth]{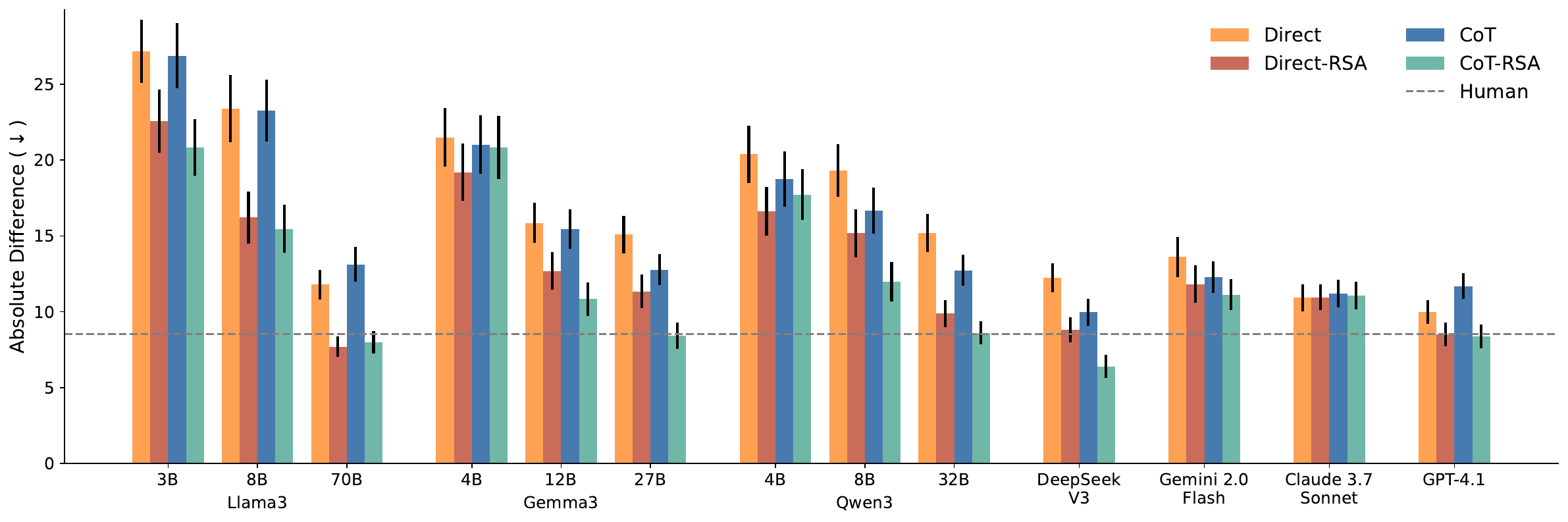}
\vspace{-0.5cm}
\caption{Absolute difference between Qwen3 32B's judgment and the target value, using clues provided by different models. Error bars show standard error over each problem. The dashed line indicates performance using human-written clues.}
\vspace{-0.2cm}
\label{fig:speaker_diff}
\end{figure*}

\section{LM Production}
\label{sec:speaker}

The language production task requires the speaker to provide a clue that best represents the target value's position on the scale. An ideal speaker, therefore, should provide an informative clue that helps the listener guess the target correctly.

\subsection{Methods}

We evaluate each LM using three methods. The \textit{direct} and \textit{zero-shot CoT} methods prompt the LM to generate the clue. The former does not explicitly encourage step-by-step reasoning, while the latter does (see prompts in Appendix~\ref{sec:app_prompts}).

The \textit{LM-RSA} method instead starts with the literal speaker $S_0$. We define \( P_{S_0}(u \mid s) \) as the LM's distribution \( P_{LM}(u \mid s) \); this distribution uses either direct or CoT prompting.
Then, the pragmatic speaker $S_1$ chooses its utterances by considering how listener might interpret their meaning, i.e. sampling the utterance from
\[
P_{S_1}(u \mid s) = \frac{P_{S_0}(u \mid s) \cdot P_{L_0}(s \mid u)}{\sum_{u' \in \mathcal{U}} P_{S_0}(u' \mid s) \cdot P_{L_0}(s \mid u')},
\]
where $P_{L_0}$ is a direct-prompted LM. This formulation can be viewed as a version of ``Inverse-RSA'' \citep{franke2022pragmatic} that starts with a pre-trained speaker and reweights candidate utterances based on pre-trained listener interpretation probabilities \cite[cf.][]{andreas2016reasoning, hendricks2016generating}. We use 32 alternative utterances in our~experiments.

\subsection{Evaluation Metrics}

Since the goal of the speaker is to provide clues that helps the listener to guess the target, we evaluate the speaker using the listener's performance, i.e., the absolute difference between the listener's prediction and the target value. We use Qwen3 32B with CoT as the listener (i.e., the judge) for most experiments~\citep{zheng2023judging}, as our results in Section~\ref{sec:comprehension} show that it achieves strong performance and exhibits high correlations with humans. 

\subsection{Results}

\begin{table}[t!]
\footnotesize
\centering
\scalebox{0.8}{
\begin{tabular}{lcccc|c}
\toprule
 & \multicolumn{5}{c}{Speaker} \\
\cmidrule(lr){2-6} 
Listener (Judge) & Direct       & Direct-RSA   & CoT          & CoT-RSA      &  Human                      \\
\midrule
Qwen3 32B & 14.39 & \textbf{7.66}  & 12.25 & \textbf{6.79}  & 9.85 \\
Llama3 70B & 15.18 & \textbf{9.87} & 12.74 & \textbf{8.62} & 8.54 \\
Human     & 15.48 & \textbf{11.92} & 14.23 & \textbf{11.61} & 8.98  \\
\bottomrule
\end{tabular}
}
\caption{Absolute difference between the prediction and the target value. We compare the performance of Qwen3 32B, Llama3 70B, and humans as listeners (judges), using clues provided by the Qwen3 32B with different methods and human speakers. The best performing methods among the four are bolded (up to statistical significance, $p <0.05$).}
\label{tab:speaker_human}
\vspace{-0.3cm}
\end{table}

\begin{table*}[t]
\footnotesize
\centering
\scalebox{0.76}{
\begin{tabular}{ccccccccccccc}
\toprule
\textbf{Left Concept} & \textbf{Right Concept} & \textbf{Target} & \textbf{Direct} & \textbf{Direct-RSA} & \textbf{CoT} & \textbf{CoT-RSA} &  \textbf{Human} \\
\midrule
Hard to remember & Easy to remember & 70 & Mnemonic \scriptsize{(60.4)} & Password tip \scriptsize{(69.8)} & Catchy song \scriptsize{(88.4)} & Song lyric \scriptsize{(62.2)} & Mindful \scriptsize{(81.4)} \\
Short & Long & 60 & Jeans \scriptsize{(84.4)} & Novella \scriptsize{(49.0)} & Poem \scriptsize{(19.6)} & Novella \scriptsize{(49.0)} & Hollywood movie \scriptsize{(59)} \\
Art & Commerce & 65 & Advertising \scriptsize{(64.6)} & Marketing design \scriptsize{(59.4)} & Advertising \scriptsize{(64.6)} & Branding \scriptsize{(74.5)} & Patreon \scriptsize{(62.2)} \\
\bottomrule
\end{tabular}
}
\caption{Qualitative examples of the LM-generated clues and human-written clues. The number in brackets shows the mean predictions from 5 human participants.}
\label{tab:qualitative_clue}
\vspace{-0.3cm}
\end{table*}

We show the performance evaluated using Qwen3 32B as the listener in Fig.~\ref{fig:speaker_diff}. Similar to findings in the language comprehension task, while smaller models perform poorly in the speaker role, larger models generally achieve strong performance. However, CoT prompting does not necessarily improve performance. By contrast, using RSA consistently improves performance over both direct and CoT prompting, bringing many LMs closer to or even better than human performance, as evaluated by the LM judge.

Note that the performance of Qwen3 32B may not be directly comparable with other models due to the potential issue of self-preference bias where the model favors its own responses~\citep{panickssery2024llm, wataoka2024self}. To eliminate this issue, we perform human evaluation on the model's produced clues by collecting 5 human listener guesses for each problem and compute their average. We also use another strong model, Llama3 70B to evaluate its performance.
We show results in Table~\ref{tab:speaker_human}. We observe similar trends using human listener and the LM listeners: RSA consistently improves over both the direct and the CoT approach, demonstrating the effectiveness of RSA in improving pragmatic language production. At the same time, on average our human-generated clues are still better than the LM-generated clues when humans are the listener (judge).

\subsection{Analysis}
\paragraph{Number of alternatives.}
One important component of RSA is reasoning about how a listener will interpret the meaning of alternative utterances. To investigate the impacts of the \textit{number} of alternative utterances, we evaluate the speaker models using different numbers of utterances. We find that increasing the number of alternatives consistently improves performance (see Appendix \ref{sec:app_ablation}).

\paragraph{Qualitative Examples.}
We show qualitative examples of the LM-generated clues (from Qwen3 32B) and human-written clues in Table~\ref{tab:qualitative_clue}. In the first example, LM-RSA methods generate intuitively and quantitatively good clues, even arguably better than human's. In the second example, both RSA methods generate the same clue that is clearly better than those of the baselines but not as good as human's. In the third example, CoT-RSA decreases performance, although all the LM-generated clues seem at least decently good. We also show example output of the Qwen3 32B speaker using CoT prompting in Appendix~\ref{sec:app_qualitative}. We find that it often reasons about a few alternatives before deciding the final ones, a process similar to what RSA framework aims to model.

\section{Discussion}

We study the pragmatic reasoning capabilities of LMs across broad concepts through the Wavenlength game. Our results show that simply using prompting,
LMs achieve strong performance on both language comprehension and production tasks, and their predictions correlate highly with human judgments.
Both task performance and humanlike-ness improve as model size increases, suggesting that larger models acquire substantial pragmatic reasoning abilities and rich conceptual knowledge---a potential benefit of scaling. Nonetheless, they also tend to diverge from human judgment distributions: whereas human judgments reflect considerable uncertainty, the LM distributions are generally more concentrated and spiky. As LMs are increasingly used as general assistants, their responses reflecting the uncertainty of the world and capturing the underlying human distribution becomes more important. Future work can investigate methods to improve humanlike reasoning capabilities---including reasoning about human beliefs and intentions---of LMs~\citep{lake2017building, collins2024building}.

Motivated by this consideration, we integrate the classical RSA framework with LMs and demonstrate its effectiveness on the language production task. This integration mitigates certain fundamental challenges inherent to the traditional RSA framework—such as defining the set of alternative utterances and specifying base-level literal listener and speaker models—which have limited its application in more realistic scenarios. It incorporates explicit probabilistic inference into LMs and can be viewed as an instance of neuro-symbolic methods \citep[cf.][]{wong2023word, puri2025probabilistic}. Our approach can improve performance not only with the direct generation setting but also when combined with CoT prompting.

Our WavelengthEval dataset offers data and human judgments to study conceptual knowledge in both humans and LMs. Future work can further extend the dataset by including more target values for each pair of concepts and investigate the conceptual representations along the scale in both humans and LMs~\citep{van2021probabilistic, grand2022semantic, tessler2022warm}. 
Such extended data with more human-written clues could also provide insights for studying human language processing. For example, when considering the ``Cheap'' and ``Expensive'' concepts, some people may reason based on their relative comparison class rather than absolute values, possibly treating ``iPhone'' as more expensive than ``Toyota'', even if their absolute costs might be the opposite~\citep{kamp1975two, cresswell1976semantics, kennedy2007vagueness, tessler2020informational}.

More philosophically, our work sheds light on the relationship between acquiring pragmatic knowledge and performing online pragmatic inference. One extreme view of pragmatics is that all of it can be learned, perhaps in a general way (e.g. through next-token prediction) such that there is nothing special about pragmatic phenomena for language acquisition. An opposite extreme view is that pragmatic interpretation and utterance are done entirely on-the-fly based on reasoning about literal meanings given context demands, and literal meanings constitute primary semantic representations in the mind. Our results suggest that there is a plausible middle ground: much of pragmatics can be learned (where literal LM listeners and speakers already perform well), but in many cases explicit reasoning is still desirable (where RSA helps). We leave more systematic development of this position for future work.

\section{Conclusion}

To sum up, our findings contribute to identifying the strengths and limitations of LM's abilities in pragmatic language comprehension and production, demonstrate the potential to improve them with RSA, and open up future opportunities to study conceptual knowledge, language use, and social reasoning in LMs and humans.

\section{Limitations}

One limitation of the present study is the size of the stimuli set, as a hundred problem instances, though typical in human studies, do not constitute a large-scale benchmark. Here we are limited by resources, given that human production and comprehension studies are expensive to run. However, since we have established that frontier models are generally good at our tasks, future studies may employ an LM-based augmentation to enlarge the dataset and evaluate smaller models. Furthermore, the overall scale of the human data we have collected is not small. For example, we have recruited over 700 human participants, and each comprehension problem has 40 human responses. We believe these human data would be valuable for experimental pragmatics and cognitive science.

Another limitation is our simplification of the Wavelength game. We formulate the game as single-turn comprehension and production tasks, whereas the real game features repeated interactions between multiple players. Learning about other players and adapting accordingly and quickly are implicit skills required to excel at the game, which our setup does not reflect, and such skills underlie real-world linguistic interactions. We believe richer settings where pragmatic communication, Theory-of-Mind, and personalization organically come together would be valuable for further evaluating and improving LMs as helpful, aligned conversational agents~\citep{liu2023improving, lin2024decision}.

A limitation on the modeling side is that, although RSA is a very general computational framework, we have only explored relatively straightforward ways of combining it with LMs. For example, RSA can include considerations about costs of utterances, which might concern an utterance's length or ease of comprehension and production. Here we have made the assumption that all utterances have the same cost. More explorations of defining and utilizing utterance costs could be interesting for future LM research.

\section*{Acknowledgments}

We thank Noah Goodman for general discussions about the RSA framework. We thank Guangyuan Jiang for helpful comments on the paper. We thank the three anonymous reviewers for their valuable feedback and constructive suggestions, which helped improve the quality of this work. This study was supported in part by funds from the MIT-IBM Watson AI Lab and a grant from the Simons Foundation to the Simons Center for the Social Brain at MIT. RPL acknowledges support from NSF grant BCS-2121074.

\bibliography{references}
\appendix

\section{Human Experiment Details}
\label{app:human_exp}

Here we include more details of human data collection.

\paragraph{Procedure.} As described in Section~\ref{sec:data}, the set of 100 problems constitutes the stimuli for our production data collection. Each participant generates a clue for a given problem, and they do so for 10 problems.
To ensure data quality we ask the participant to think for at least 20 seconds before generating the clue.
We collect 5 clues for each problem and 500 clues in total. 
Participants' median completion time for the generation task is approximately 9 minutes. 
To potentially increase clue diversity and quality, the author team also composes one clue for each problem. We combine these clues with participant-composed clues for the next stage---filtering out the best clue for each problem, for which we collect 15 human judgments (guessing the target value given the corresponding clue using a slider). We keep the best clues based on mean human absolute differences from the target value, finalizing a set of 100 problems for the official comprehension task.

In the comprehension task, similarly to ensure data quality we ask the participant to think for at least 10 seconds before guessing the target value using a slider. Each participant makes guesses for approximately 20 problems, which has a median completion time of 8 minutes. On each comprehension problem, we collect 25 judgments, and plus the previous 15 judgments we have total 40 human judgments. This relatively large number of judgments allows us to have a more representative distribution of human uncertain judgments.

\begin{figure*}[htp]
    \centering
    \includegraphics[width=0.95\linewidth]{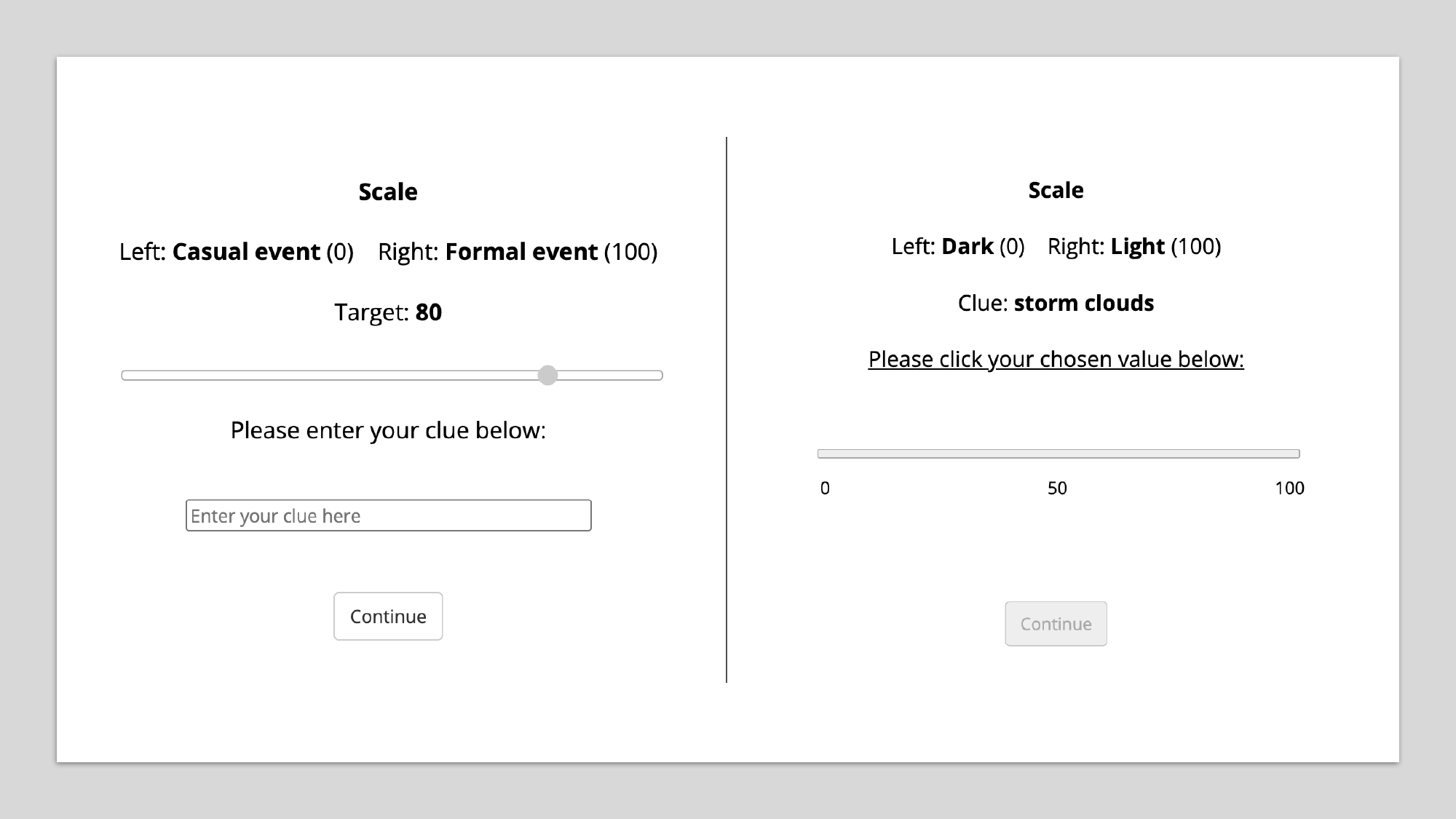}
    \caption{Experimental interface for the human studies. One the left and right show the production (clue generation) and comprehension (target guessing) interfaces, respectively.}
    \label{fig:human_exp}
\end{figure*}

\paragraph{Instructions and interface.} The main interface for the human experiments is shown in Fig.~\ref{fig:human_exp}. The verbal instructions we provide to human participants are highly similar to the listener and speaker prompts we provide to LMs (shown below in Section \ref{sec:app_prompts}).

\paragraph{Cost.} Human participants on average are paid at a rate of at least \$15 per hour. The production (clue generation) human experiment costs $\sim$\$120. The comprehension human experiment (including filtering for good clues) costs $\sim$\$1000. The human evaluation of model generated clues costs $\sim$\$150. So the overall cost is $\sim$\$1270.

\section{Language Model Details}

All open-weights models except Deekseek-V3 are run on local H100 and A100 clusters, and all corresponding experiments can be finished in 12 hours with 4 H100s. Deepseek-V3 is run on the \href{http://www.together.ai}{Together AI} inference platform via APIs. Close models Gemini 2.0 Flash (\texttt{001}), Claude 3.7 Sonnet (\texttt{20250219}), and GPT-4.1 (\texttt{2025-04-14}) are called using their respective APIs. All API calls combined cost under \$500.

\section{Alternative RSA Listener Model}
\label{sec:app_listener_v2}

We explore an alternative RSA model for the LM comprehension task. Specifically, we define the pragmatic listener $L_1$ as 
\[
P_{L_1}(s \mid u) = \frac{P_{L_0}(s \mid u) \cdot P_{S_0}(u \mid s)}{\sum_{s' \in \mathcal{S}} P_{L_0}(s' \mid u) \cdot P_{S_0}(u \mid s')}.
\]
This model is similar to our pragmatic speaker  $S_1$ in Section~\ref{sec:speaker}, except that we marginalize over all states $\mathcal{S}$ instead of all utterances $\mathcal{U}$. We use the length-normalized probability of an directly prompted LM as an unnormalized estimate of $P_{S_0}(u \mid s)$. Since this requires having access to the model's next-token probability, we only evaluate this using open-weights models. 

\begin{figure*}[htp]
    \centering
    \includegraphics[width=0.85\linewidth]{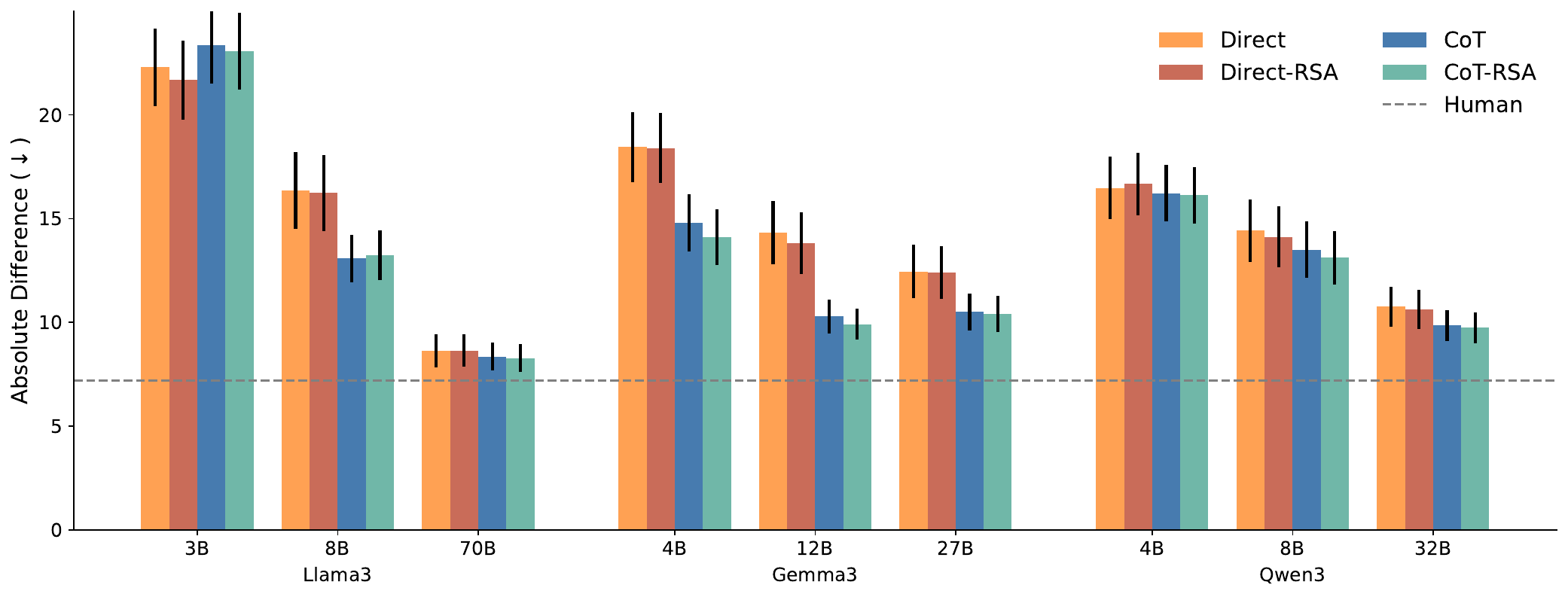}
    \caption{Absolute difference between the model's prediction and the target value. We use the alternative RSA model for Direct-RSA and CoT-RSA. Error bars show standard error over each problem. The dashed line indicates human performance.}
    \label{fig:listener_diff_v2}
\end{figure*}

\begin{figure*}[htp]
    \centering
    \includegraphics[width=0.85\linewidth]{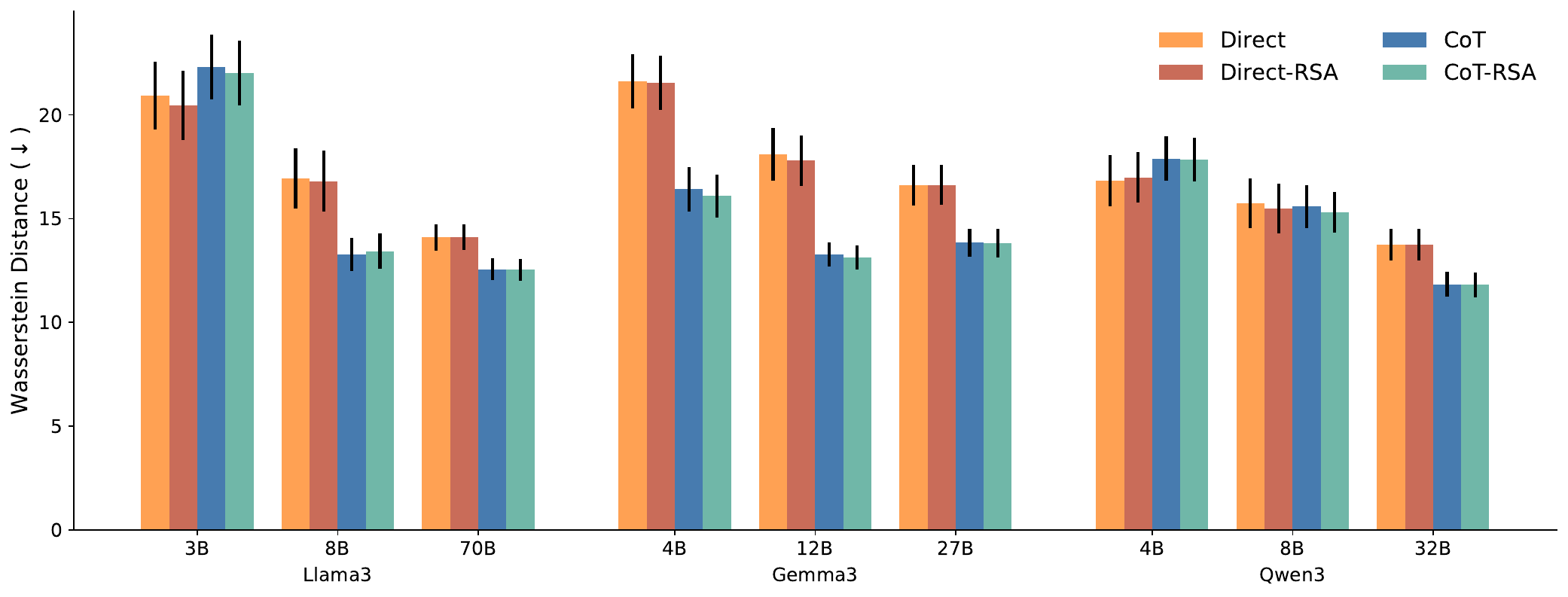}
    \caption{The Wasserstein distance between the model distribution and human distribution. We use the alternative RSA model for Direct-RSA and CoT-RSA. Error bars show standard error over each problem.}
    \label{fig:listener_human_dist_v2}
\end{figure*}

We show the results in Fig.~\ref{fig:listener_diff_v2} and Fig.~\ref{fig:listener_human_dist_v2}. We find that this alternative performs similarly to the RSA model described in Section~\ref{sec:comprehension}, and incorporating RSA does not necessarily improve task performance or human-likeness on the comprehension task.

\section{Additional LM Results}
\label{sec:app_results}

\subsection{Listener Distribution Entropy}

\begin{figure*}[htp]
    \centering
    \includegraphics[width=\linewidth]{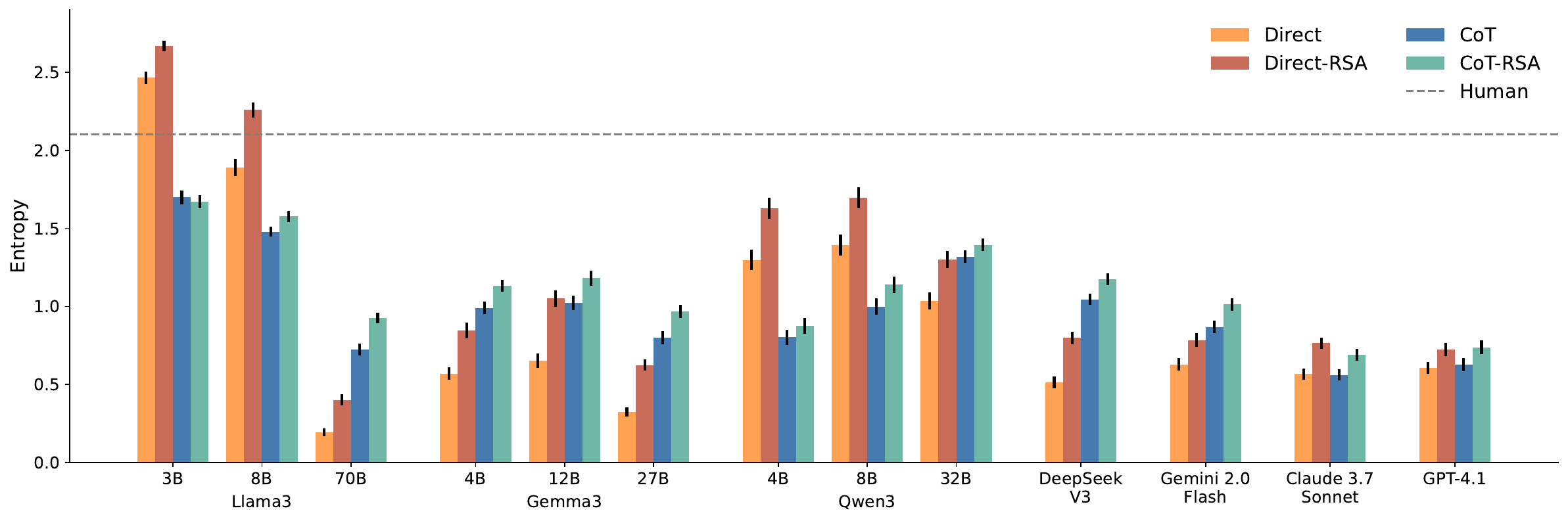}
    \caption{Entropy of the LM listener distribution. Error bars show standard error over each problem. The dashed line indicates human performance.}
    \label{fig:listener_entropy}
\end{figure*}

We measure the entropy of the listener distribution and show results in Fig.~\ref{fig:listener_entropy}. We find that the LM distributions generally have lower entropy than that of human distributions, indicating that human distributions reflect more uncertainty. We observe that RSA almost always increases entropy over the base prompting method.

\subsection{Comprehension Task Performance Breakdown}

\begin{figure*}[t]
    \centering
    \includegraphics[width=\linewidth]{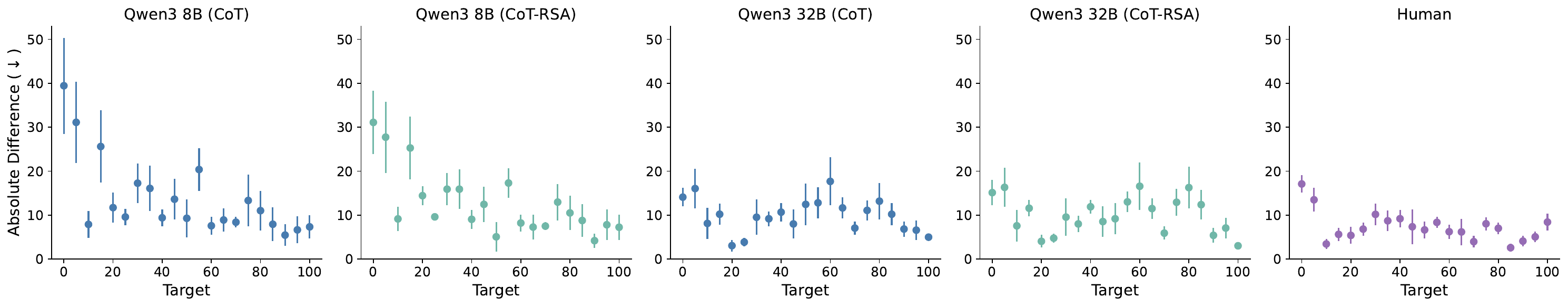}
    \caption{The average absolute difference between the model or human prediction and the target, broken down by target value. Humans perform worse at the extreme scale (0), but are generally consistent across different target values. The performance of LMs, however, shows larger variance across target values, especially for small model (Qwen3 8B) which performs significantly poorly near 0. Error bars show standard error over each target value.}
    \label{fig:target_vs_diffs}
\end{figure*}

While our target values uniformly range from 0 to 100, some might be easier to guess. In Fig.~\ref{fig:target_vs_diffs}, we show the average absolute difference broken down by target value. We find that the smaller LM (Qwen3 8B) has larger variance across target values, whereas humans and the larger LM (Qwen3 32B) are more consistent.

\subsection{Choices of Alternatives in the Comprehension Task}
\label{sec:app_listener_alt}
\begin{table*}[t!]
\footnotesize
\centering
\scalebox{1.0}{
\begin{tabular}{lllcccc}
\toprule
& & & \multicolumn{2}{c}{Absolute Difference ($\downarrow$)} & \multicolumn{2}{c}{Wasserstein Distance ($\downarrow$)} \\
\cmidrule(lr){4-5}  \cmidrule(lr){6-7}  
Model & Listener & Speaker  & Direct-RSA & CoT-RSA & Direct-RSA & CoT-RSA \\
\midrule
\multirow{4}{*}{Llama3} & 3B & 3B & 22.49 & 21.63 & 20.71 & 21.13 \\
& 3B & 70B & 22.05 & 21.79 & 20.39 & 20.69 \\
& 8B & 8B & 15.49 & 12.89 & 15.44 & 12.30 \\
& 8B & 70B & 16.16 & 12.96 & 16.01 & 12.65 \\
\midrule
\multirow{7}{*}{Gemma3} & 4B & 4B & 18.25 & 14.48 & 20.90 & 16.25 \\
& 4B & 27B & 17.84 & 14.49 & 20.35 & 16.10 \\
& 12B & 12B & 13.77 & 10.35 & 17.13 & 12.88 \\
& 12B & 27B & 13.46 & 10.45 & 16.49 & 12.89 \\
& 27B & 4B & 12.14 & 10.06 & 16.09 & 13.40 \\
& 27B & 12B & 12.02 & 10.82 & 15.98 & 13.71 \\
& 27B & 27B & 12.07 & 10.51 & 15.87 & 13.11 \\
\midrule
\multirow{7}{*}{Qwen3} &  4B &  4B & 17.50 & 16.70 & 17.42 & 18.22 \\
& 4B &  32B & 16.97 & 15.76 & 16.91 & 17.54 \\
& 8B &  8B & 13.37 & 12.65 & 14.36 & 14.22 \\
& 8B &  32B & 13.82 & 13.19 & 14.50 & 15.02 \\
& 32B &  4B & 11.30 & 9.93 & 13.53 & 11.79 \\
& 32B &  8B & 10.82 & 9.95 & 13.24 & 11.94 \\
& 32B &  32B & 11.00 & 10.05 & 13.25 & 12.02 \\
\bottomrule
\end{tabular}
}
\caption{Absolute difference between the model's prediction and the target value, and the Wasserstein distance between the model distribution and the human distribution. We compare the model's performance under two conditions: using alternatives generated by the model itself versus those generated by a different LM.}
\label{tab:alternatives}
\end{table*}

We do not observe improvements when incorporating the RSA model for the comprehension task. Here, we investigate two possible hypotheses for why it does not provide further improvement: (1) the LM listener distribution is concentrated and spiky, which limits the benefits of RSA, or (2) the LM does not generate good alternatives as a simulated speaker.

To investigate which factor contributes more to the performance, we provide each listener LM with alternatives generated from a different speaker LM. Since the stronger LMs tend to generate better alternatives (as evidenced by their better performance on the generation task in Fig.~\ref{fig:speaker_diff}), if the weaker model benefits from better alternatives, this would suggest its performance is limited by its ability to generate good alternatives. Similarly, if a stronger model's performance degrades when using alternatives from a weaker LM, it would confirm that the ability to propose good alternatives is a key factor. In contrast, if there is no significant performance difference when an LM uses its own alternatives versus those from another model, this would suggest that the LM's listener distribution may be the main issue.

We show results in Table~\ref{tab:alternatives}. For most models, we do not observe a significant difference between the original results and the ones using alternatives from another model, which suggests that for the comprehension task, the performance is less sensitive to the choices of alternatives. Therefore, we hypothesize that the LM's listener distribution may be a more important limiting factor.

\subsection{Number of Alternatives in the Production Task}
\label{sec:app_ablation}

\begin{figure}[t!]
    \centering
    \includegraphics[width=\linewidth]{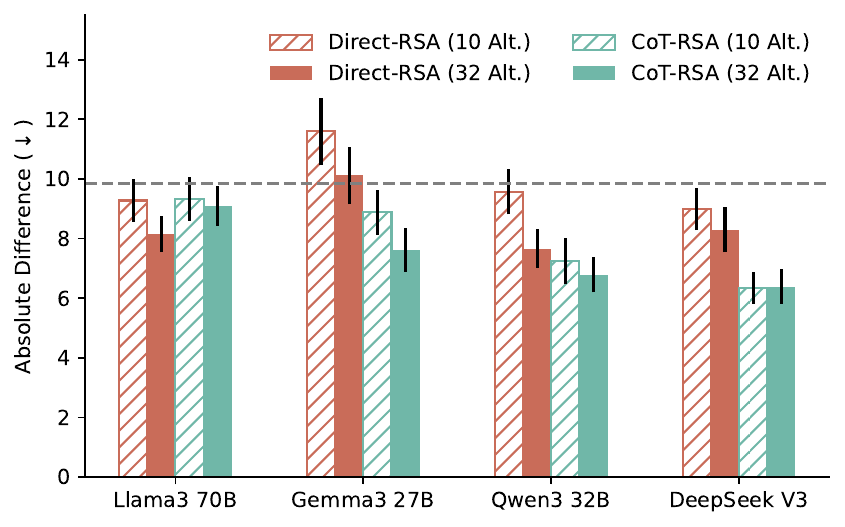}
    \caption{Ablations of the number of alternatives for the language production task. We compare the RSA-based LM speakers using 10 or 32 alternatives. The results show the absolute difference between the listener model's (Qwen3 32B) prediction and the target value.}
    \label{fig:speaker_alt}
\end{figure}

In the LM production task setting, we observe significant improvement using RSA on top of both direct and CoT prompting. One important component of RSA is reasoning about how a listener will interpret the meaning of alternative utterances. To investigate the impacts of the number of utterances, we evaluate the speaker models but use different numbers of utterances. As shown in Fig.~\ref{fig:speaker_alt}, across the best-performing models from different families, increasing the number of alternatives consistently improves performance.

\subsection{Full Results}
\label{sec:app_full_results}

\begin{figure*}[t]
    \centering
    \includegraphics[width=\linewidth]{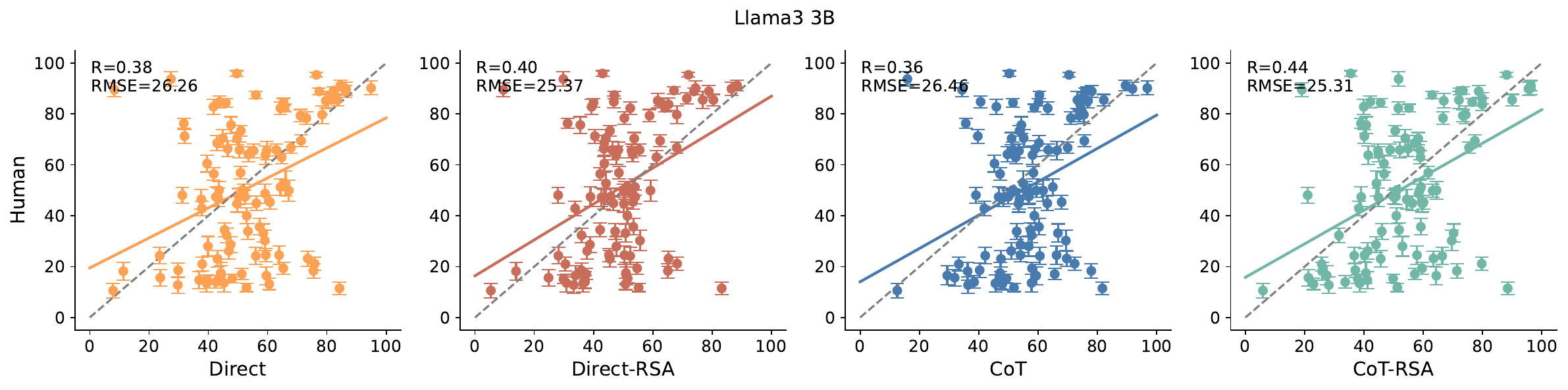}
    \includegraphics[width=\linewidth]{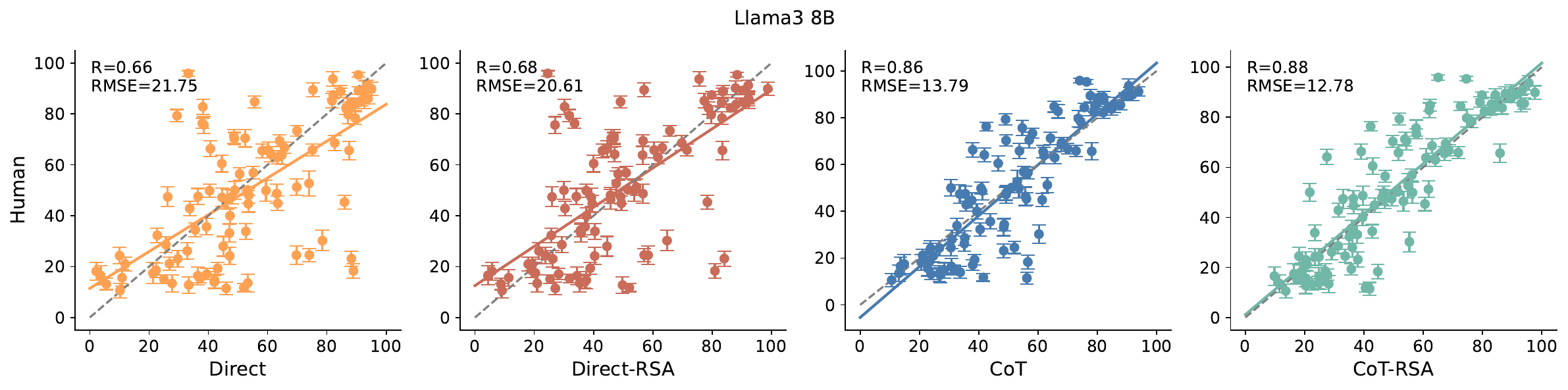}
    \includegraphics[width=\linewidth]{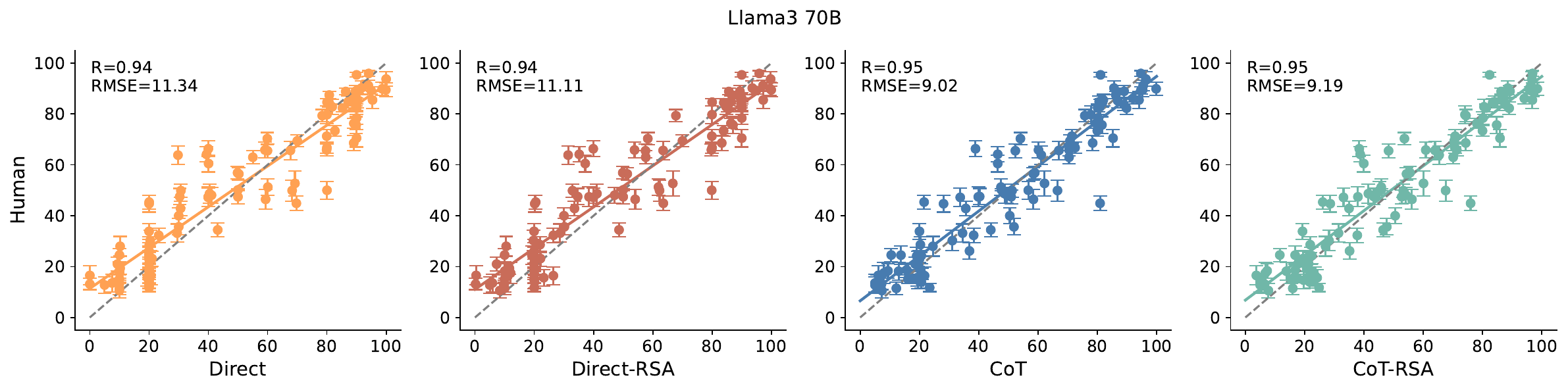}
    \caption{Correlations between model predictions and human judgments using Llama3 models. We show Pearson correlations and root mean square standard error (RMSE). Error bars show standard error over 40 human participants.}
    \label{fig:corr_llama3}
\end{figure*}

\begin{figure*}[t]
    \centering
    \includegraphics[width=\linewidth]{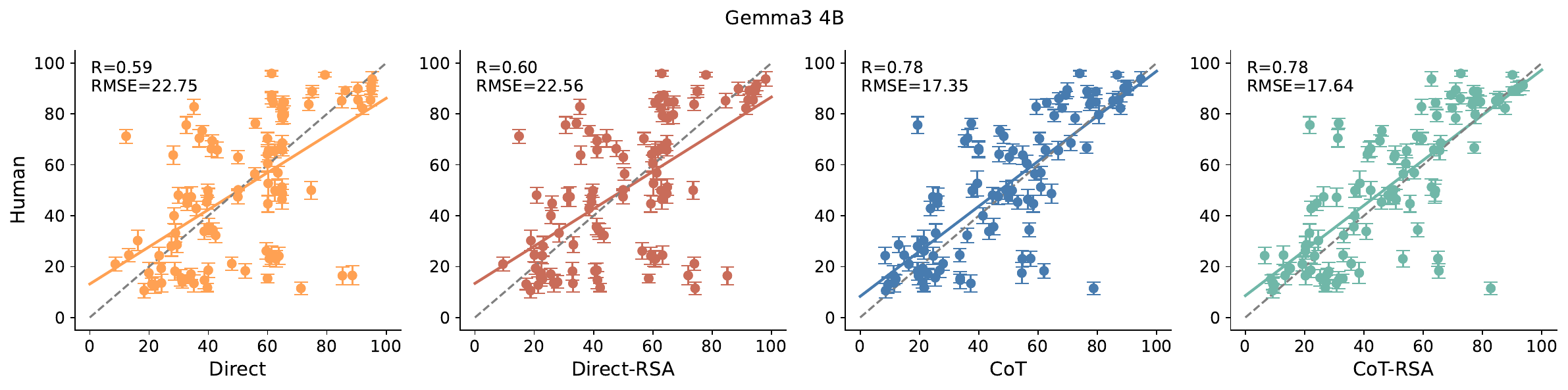}
    \includegraphics[width=\linewidth]{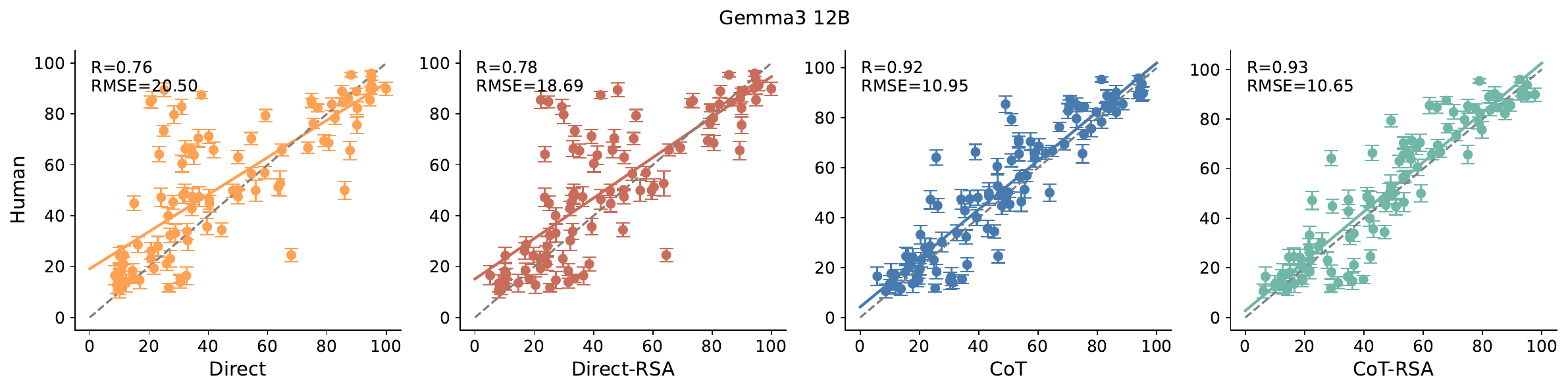}
    \includegraphics[width=\linewidth]{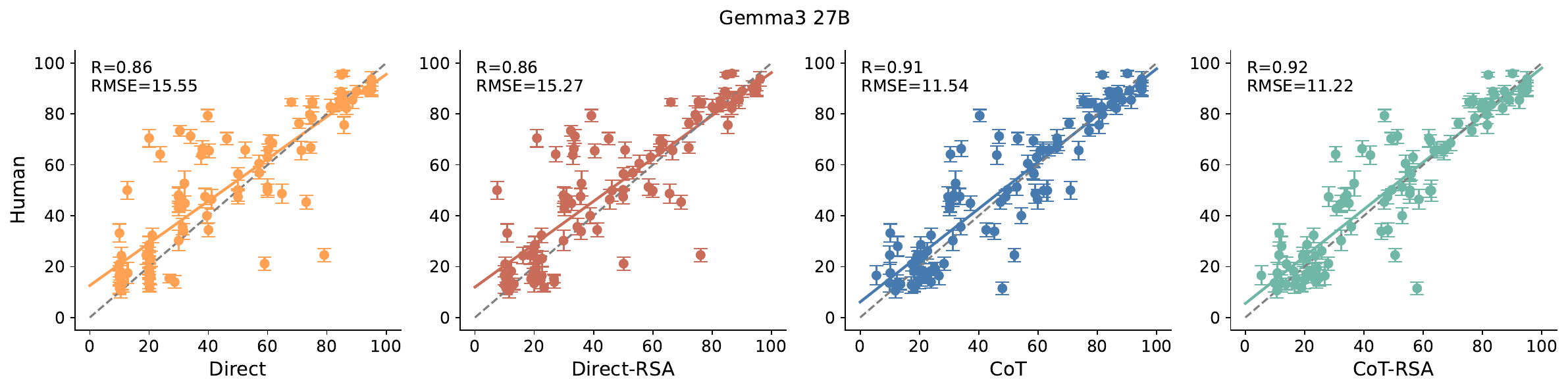}
    \caption{Correlations between model predictions and human judgments using Gemma3 models. We show Pearson correlations and root mean square standard error (RMSE). Error bars show standard error over 40 human participants.}
    \label{fig:corr_gemma3}
\end{figure*}

\begin{figure*}[t]
    \centering
    \includegraphics[width=\linewidth]{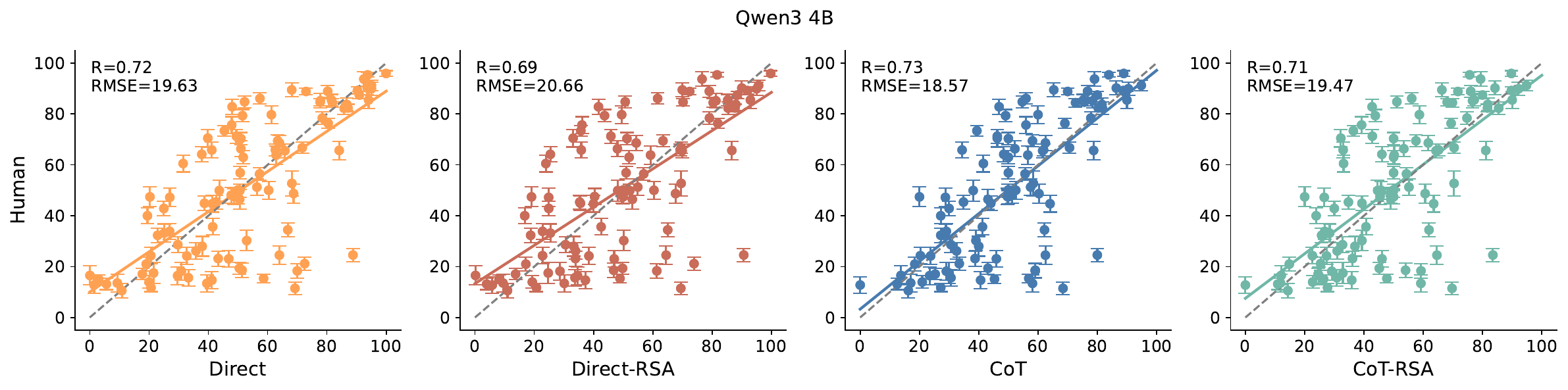}
    \includegraphics[width=\linewidth]{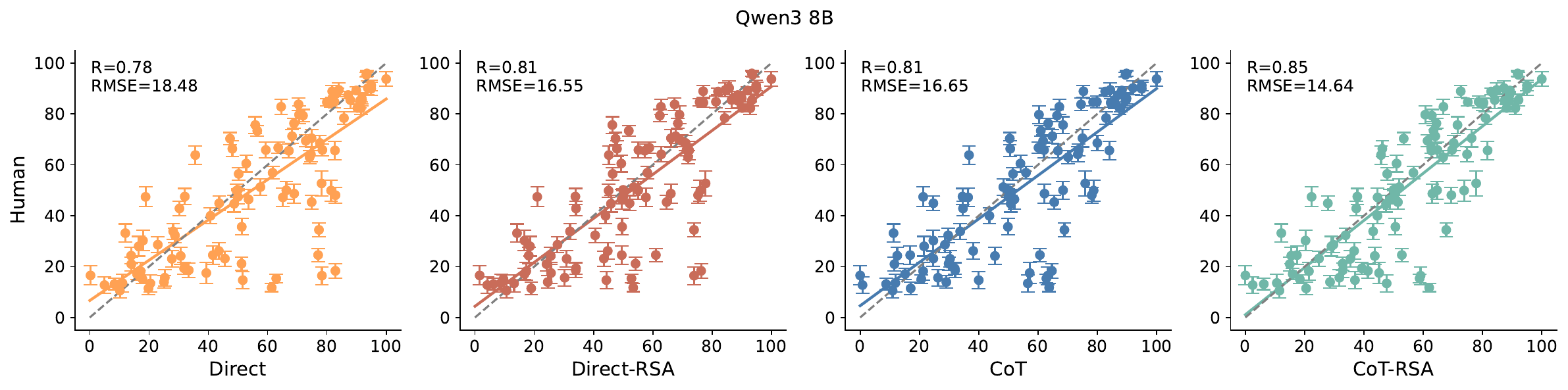}
    \includegraphics[width=\linewidth]{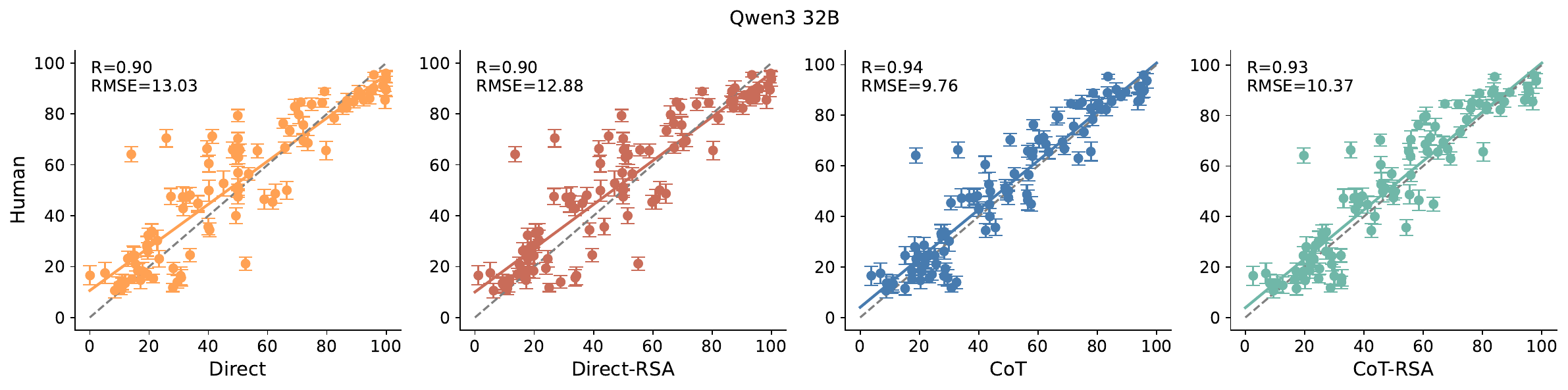}
    \caption{Correlations between model predictions and human judgments using Qwen3 models. We show Pearson correlations and root mean square standard error (RMSE). Error bars show standard error over 40 human participants.}
    \label{fig:corr_qwen3}
\end{figure*}

\begin{figure*}[t]
    \centering
    \includegraphics[width=\linewidth]{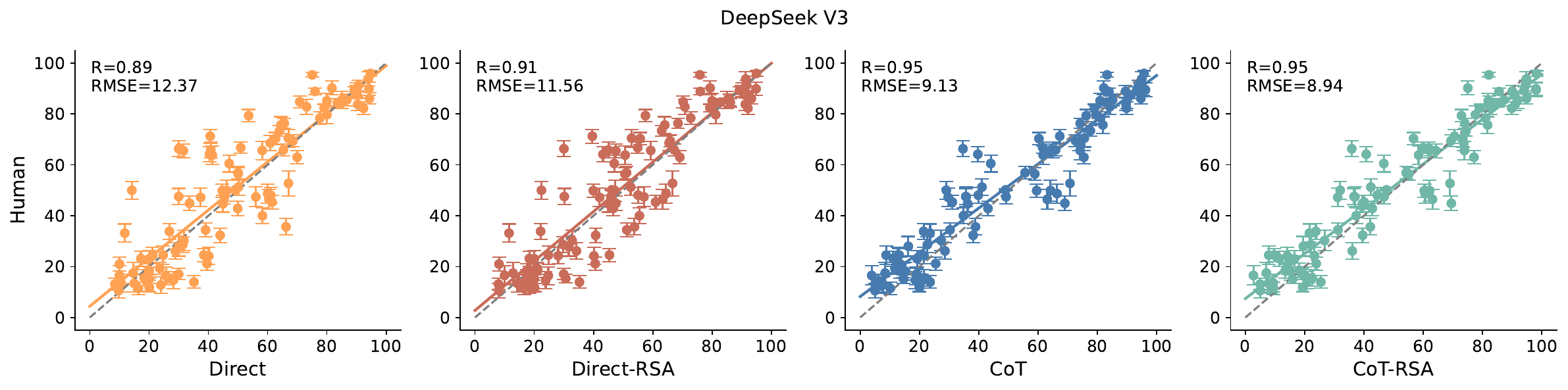}
    \includegraphics[width=\linewidth]{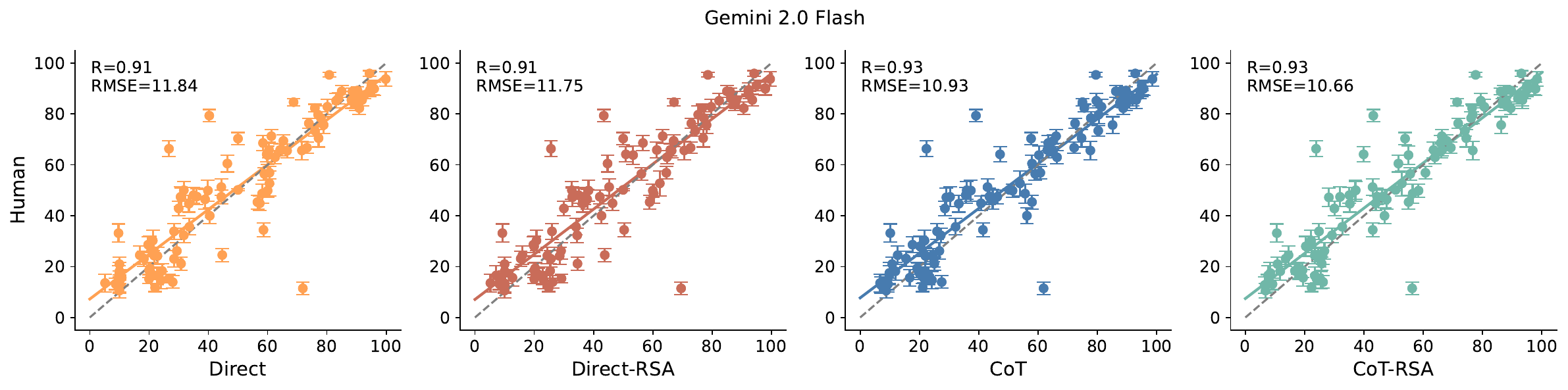}
    \includegraphics[width=\linewidth]{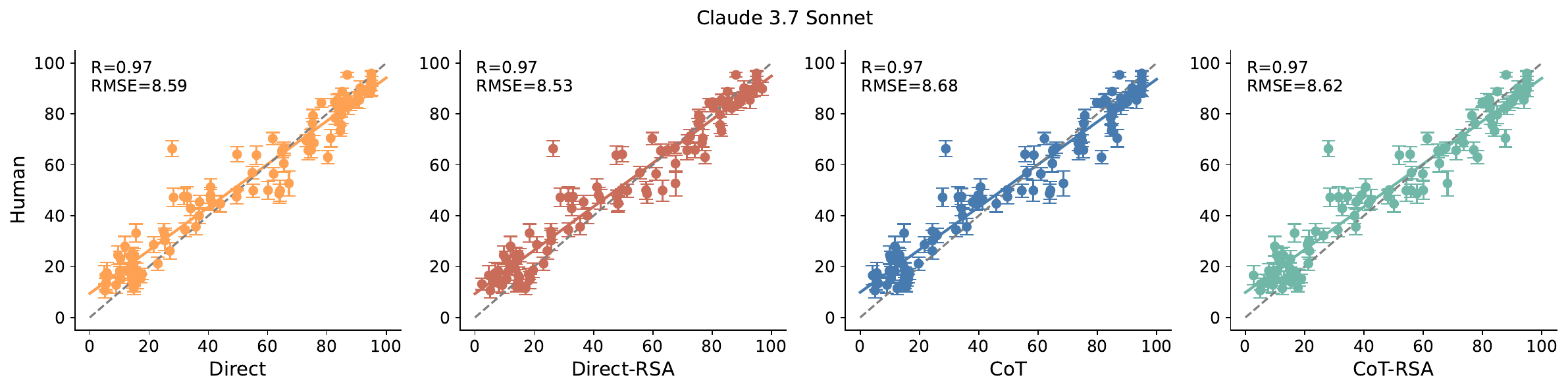}
    \includegraphics[width=\linewidth]{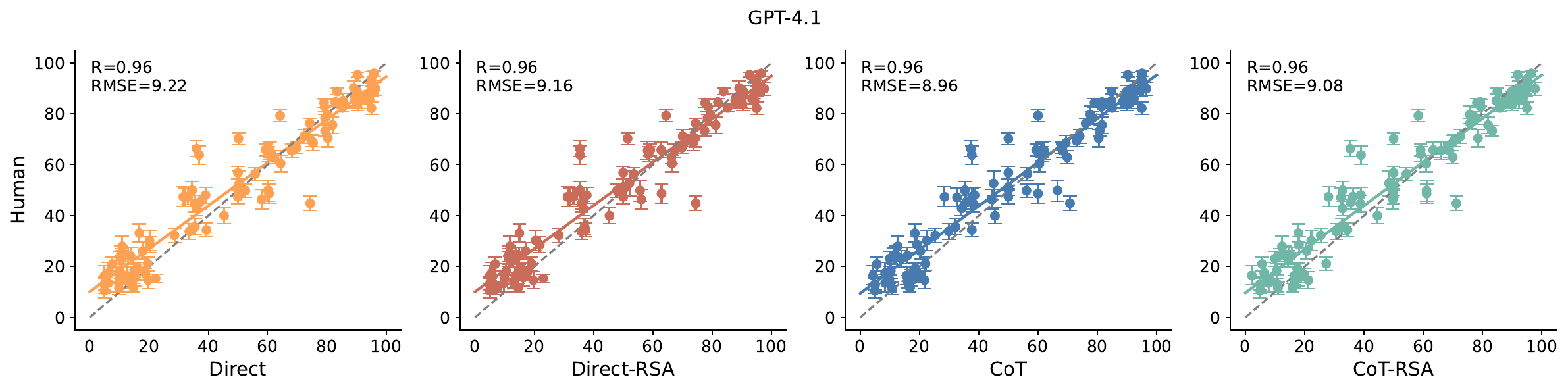}
    \caption{Correlations between model predictions and human judgments using DeepSeek-V3 and closed-source models. We show Pearson correlations and root mean square standard error (RMSE). Error bars show standard error over 40 human participants.}
    \label{fig:corr_other}
\end{figure*}

We show the correlations between model predictions and human judgments in Fig.~\ref{fig:corr_llama3}--Fig.~\ref{fig:corr_other}.

\newpage
\section{Qualitative Example Outputs}
\label{sec:app_qualitative}

The following is an example output of Qwen3 32B listener using CoT prompting.

\begin{tcbverbatim}[Qwen3 32B listener output using CoT]
{basicstyle=\ttfamily\scriptsize, label={tab:listener_ex}}
To determine where "soup" falls on the scale between Snack (0) and Meal (100), we can consider whether soup is generally associated more with a light snack or a full meal. 

- **Snack (0):** Snacks are typically small, quick, and meant to satisfy hunger briefly. Examples include chips, fruits, or a granola bar.
- **Meal (100):** Meals are more substantial, often part of a regular eating pattern (like breakfast, lunch, or dinner), and may include multiple components.

- **Soup as a clue:** Soup can be both a snack and part of a meal, depending on the context. For example, a small can of soup eaten on the go might be a snack, but a bowl of soup served with a salad and bread could be part of a full lunch. Soup is often served as an appetizer or a main dish in a meal.

Given this flexibility, but leaning toward being a component of a meal rather than a simple snack, a reasonable estimate for "soup" on the Snack-Meal scale would fall around the **middle to slightly higher** part of the spectrum.

<answer>60</answer>
\end{tcbverbatim}

The following is an example output of Qwen3 32B speaker using CoT prompting.

\begin{tcbverbatim}[Qwen3 32B speaker output using CoT]
{basicstyle=\ttfamily\scriptsize}
The target value is 70 on a scale from Colorless (0) to Colorful (100), which means the clue should represent something that is **fairly colorful**, but not **extremely** vibrant or multicolored. It should be on the **brighter end** of the spectrum, but still allow room for something more intensely colorful to exist. A strong clue would evoke something that has **distinct color**, but may not have **a wide variety of colors**.

Good clues might include: 
- **Sunset** - vibrant but not multicolored in the extreme  
- **Tropical fish** - colorful but still short of the most dazzling hues  
- **Rainbow flags** - colorful but not the most vivid version of color (e.g., lasers, stained glass)  
- **Flowers** - can be colorful, but the term is general enough to allow for variation  
- **Marigold** - a specific, strong color but still within the colorful range  

Clues that would be **too colorless** might include:  
- Gray (closer to 0)  
- Concrete (very near to 0)  

Clues that would be **too colorful** might include:  
- Neon (closer to 100)  
- Stained glass (also closer to 100)  

Therefore, a well-chosen clue for 70 might be:  

<answer>sunrise</answer>
\end{tcbverbatim}

\newpage

\section{Prompt Details}
\label{sec:app_prompts}
The following are our prompts for eliciting listener judgments and speaker productions from LMs.

\begin{tcbverbatim}[Listener prompt]
{basicstyle=\ttfamily\scriptsize}
# Context

You will be playing a language game. In each round, you will be presented a scale from 0 (left) to 100 (right). Each scale represents a spectrum between two opposite or contrasting concepts. One concept will be on the left side of the scale (towards 0), and the other concept will be on the right side (towards 100).

For each scale, you will see a clue-a single word or a short phrase-that someone else has provided. Your task is to guess a value between 0 and 100 (in increments of 5) that you believe best represents where that clue falls on the scale between the two concepts. Essentially, you are trying to interpret where on the spectrum the clue giver intended their clue to represent. A good guess should be close to the target value that the clue giver was trying to communicate. (The clue giver chose the clue with respect to the value provided to them.)

Let's look at an example:

Imagine the scale is from Slow (0) to Fast (100).

If the clue is 'race car', you would likely guess a high value, perhaps around 85, since a race car represents something that is very fast, but not the absolute fastest thing possible. If the clue had been 'turtle' instead, you might guess a value closer to 10, or if it had been 'bicycle', you might guess a value closer to 50 on this particular scale.

Here's another example:

Scale: Summer (0) to Winter (100). 
Clue: ice cream.

This clue is more strongly associated with 'Summer' than with 'Winter'. A reasonable guess might be around 30. While people certainly eat ice cream year-round, it is especially popular during hot summer days. If the clue had been 'wearing a shirt' you might guess a value closer to 50, or if it had been 'snow' you might guess a value closer to 90 or 100 in this context.

Here're a few more examples:

Scale: Feels bad (0) to Feels good (100).
Clue: relaxing.
<answer>70</answer>

Scale: Hard to spell (0) to Easy to spell (100).
Clue: Daenerys Targaryen.
<answer>20</answer>

Scale: Red (0) to Yellow (100).
Clue: tangerine.
<answer>50</answer>

Provide your best estimate carefully (in increments of 5, so the possible values are 0, 5, 10, 15, 20, 25, 30, 35, 40, 45, 50, 55, 60, 65, 70, 75, 80, 85, 90, 95, 100). The goal is to get as close as possible to the target value that the clue giver intended to communicate. If you find a clue confusing, meaningless, or hard to understand, please still make an educated guess. Some clues may not be very good, but please always make an honest attempt.

Format your response as:

<answer>your guess</answer>

# Problem

Scale: {left_word} (0) to {right_word} (100).
Clue: {clue}.
\end{tcbverbatim}

\begin{tcbverbatim}[Speaker prompt]
{basicstyle=\ttfamily\scriptsize}
# Context

You will be playing a language game. In each round, you will be presented a scale from 0 (left) to 100 (right). Each scale represents a spectrum between two opposite or contrasting concepts. One concept will be on the left side of the scale (towards 0), and the other concept will be on the right side (towards 100).

For each scale, you will see a target value, indicated by a number between 0 and 100. Your task is to think of a clue (often times a single word, or a very short phrase if necessary) that you believe best represents that specific target value's position on the scale, considering the two concepts. Essentially, you are trying to communicate where the target value lies on the spectrum using just your clue. A good clue should allow another person to approximately guess what the target value is.

Let's look at an example:

Imagine the scale is from Slow (0) to Fast (100).

If the target value is 85, you need a clue that represents something that is very fast, but perhaps not the absolute fastest. Good potential clues might include: 'professional sprinters', 'leopard', or 'Ferarri'. A clue like 'turtle' (which might be closer to 0) or 'bicycle' (which might be closer to 50) would be less appropriate for the target of 85 on this particular scale.

Here's another example:

Scale: Summer (0) to Winter (100).

Target value: 30. This target is closer to 'Summer' than to 'Winter'. Good clues might be: 'swimming', 'ice cream', or 'sunscreen'. Clues like 'wearing a shirt' (arguably closer to 50) or 'snow' (arguably closer to 100) would be less fitting for a target of 30 in this context.

Here're a few more examples:

Scale: Feels bad (0) to Feels good (100).
Target value: 70.
<answer>relaxing</answer>

Scale: Hard to spell (0) to Easy to spell (100).
Target value: 20.
<answer>Daenerys Targaryen</answer>

Scale: Red (0) to Yellow (100).
Target value: 50.
<answer>tangerine</answer>

You are encouraged to be concise when you come up with the clue. Try using a single word or a short phrase (a few words). You are not allowed to use more than 5 words for a clue. Your clue should convey a single thought. Do not combine multiple ideas. Do not use words that already appear in the two concepts or any synonyms (e.g., 'Piece' is not allowed for the 'Peaceful - Warlike' scale.) Do not use modifiers (words like 'but', 'very', 'almost', and 'slightly'). Do not use specific numbers or numeric values (including time, percentage, etc.). Do not use modifiers (words like 'but', 'very', 'almost', and 'slightly'). Do not use specific numbers or numeric values (including time, percentage, etc.).

Provide the best clue you can think of. The best clues maximize the chance that an average person can approximately guess the target value given the clue.

Format your response as:

<answer>your clue</answer>

# Problem

Scale: {left_prompt} (0) to {right_prompt} (100).
Target value: {target_value}.
\end{tcbverbatim}

\section{Concept Pairs}
\label{sec:concept_pairs}

Here is a table presenting all the concept pairs used as stimuli in our study.

\begin{table}[h!]
\centering
\scalebox{0.85}{
\begin{tabular}{ccc}
\toprule
Index & Left Concept & Right Concept \\
\midrule
1  & Bad                 & Good               \\
2  & Hot                 & Cold               \\
3  & Colorless           & Colorful           \\
4  & Low calorie         & High calorie       \\
5  & Inessential         & Essential          \\
6  & Cheap               & Expensive          \\
7  & Rare                & Common             \\
8  & Difficult to use    & Easy to use        \\
9  & Worst day of the year & Best day of the year \\
10 & Bad habit           & Good habit         \\
11 & Dark                & Light              \\
12 & Hard to remember    & Easy to remember   \\
13 & Unhealthy           & Healthy            \\
14 & Normal pet          & Exotic pet         \\
15 & Happens slowly      & Happens suddenly   \\
16 & Mental activity     & Physical activity  \\
17 & Need                & Want               \\
18 & Dry food            & Wet food           \\
19 & Optional            & Mandatory          \\
20 & Hard to pronounce   & Easy to pronounce  \\
21 & Low quality         & High quality       \\
22 & Plain               & Fancy              \\
23 & Quiet place         & Loud place         \\
24 & Dangerous           & Safe               \\
25 & Useless major       & Useful major       \\
26 & Bad for you         & Good for you       \\
27 & Waste of time       & Good use of time   \\
28 & Nobody does it      & Everybody does it  \\
29 & Snack               & Meal               \\
30 & Soft                & Hard               \\
31 & Square              & Round              \\
32 & Temporary           & Permanent          \\
33 & Sport               & Game               \\
34 & Messy food          & Clean food         \\
35 & Vice                & Virtue             \\
36 & Unpopular activity  & Popular activity   \\
37 & Boring              & Exciting           \\
38 & Easy to do          & Hard to do         \\
39 & Nature              & Nurture            \\
40 & Limited             & Infinite           \\
41 & Casual event        & Formal event       \\
42 & Small talk          & Heavy topic        \\
43 & Short               & Long               \\
44 & Talent              & Skill              \\
45 & Unnatural           & Natural            \\
46 & Funny topic         & Serious topic      \\
47 & Not enough          & Too much           \\
48 & Art                 & Commerce           \\
49 & Deep thought        & Shallow thought    \\
50 & Blue                & Green              \\
\bottomrule
\end{tabular}}
\caption{All fifty pairs of concepts used as the experimental stimuli.}
\end{table}

\end{document}